# Evaluating Nonuniform Dependability Across Response Conditions: A Conditional Generalizability Framework Illustrated in Automated Essay Scoring

Yi Gui
Measurement Incorporated
queenm@gmail.com

**Abstract**

Aggregate reliability and dependability estimates can obscure heterogeneity in measurement-design burden across response conditions; when this occurs, a single G- or D-study may mischaracterize the adequacy of a design for particular strata. This study introduces a conditional generalizability framework with three components. First, automated scoring configurations—the encoder architectures and scoring-head families admissible within a fixed development pipeline—are conceptualized as a universe of admissible measurement conditions rather than as incidental modeling choices. Second, analytical D-study projections are compared with empirical configuration sweeps over a finite scoring pool, yielding two estimands of design adequacy whose agreement or divergence diagnoses the realized configuration universe. Third, G- and D-study evidence is conditioned on entropy-defined response strata, treating entropy as an operational stratification variable, not as a construct claim about writing quality, so that both the aggregate and condition-specific design burden can be evaluated. Whereas recent extensions of generalizability theory have addressed AI-generated item variants on the response side, the present framework addresses an analogous scoring-side problem: AI-mediated scoring configurations. The framework is demonstrated using automated essay scoring of timed L2 writing tasks. The realized scoring design was dependable in aggregate ($\Phi \approx .76$ for the person × task × configuration design). Re-estimated at the essay level within entropy strata, dependability was uniformly high but declined modestly and robustly across strata ($\Phi = .88, .87, .84$)—a small gradient that nonetheless implied different decision-study requirements, with the highest-entropy stratum needing more crossed scoring conditions to reach a given dependability target. Empirical configuration sweeps further indicate whether analytical minimum-design recommendations are attainable within the available scoring pool. The framework provides a portable workflow for evaluating nonuniform dependability whenever multiple scoring, rating, or observation configurations are available.

## Introduction

Aggregate reliability and dependability estimates can obscure heterogeneity in measurement-design burden across response conditions. A single generalizability or decision study summarizes how much variability would arise under a stipulated universe of generalization, but it implicitly assumes that this burden is sufficiently homogeneous across persons, tasks, and the response conditions a scoring procedure encounters. When that homogeneity fails, a nominally adequate scoring design may be insufficient for some strata of responses and inefficient for others, even when aggregate evidence is reassuring.

This question has acquired new operational urgency in measurement settings where artificial intelligence participates in the design itself. When scores come from an automated pipeline, the conditions of measurement extend beyond prompts, occasions, and human raters to include the scoring configuration: the text encoder that represents the response and the scoring head that maps that representation to a score, among other implementation choices a pipeline may treat as replicable (such as random seeds, prompt designs, or calibration procedures). Two configurations built from the same data and pipeline can assign the same response different scores, so an examinee's reported score depends partly on which configuration a developer happened to select. The automated scoring configuration is itself a score-generating condition, and the population of such conditions can be enumerated, sampled, and replicated.

Generalizability theory provides the error-source accounting required to evaluate measurement designs of this kind (Cronbach, Gleser, Nanda, & Rajaratnam, 1972; Brennan, 2001; Webb, Shavelson, & Haertel, 2006). It defines a universe of admissible observations, decomposes observed-score variance into components for the object of measurement, the design facets, and their interactions, and uses decision studies to project relative generalizability and absolute dependability under alternative procedures. Recent work has begun to extend this framework to settings in which AI participates in the design itself (Lee, Kim, & Shin, 2026), but these extensions have so far addressed AI on the response side—treating items as randomly parallel forms generated from templates or AI systems. The analogous problem on the scoring side has not received the same attention: the response is fixed, but the automated scoring system is one realization sampled from a universe of admissible configurations.

A second methodological concern is whether dependability, however estimated, is uniform across the responses being scored. The measurement standards make the conditional-precision principle explicit, requiring that conditional standard errors of measurement be reported across score levels and across subgroups where uniformity cannot be defended (AERA, APA, & NCME, 2014, Standards 2.14–2.15). This principle motivates a parallel question for AI-era automated scoring: whether precision should also be evaluated across regions of the response space, not only across score levels. Essays differ systematically in lexical diversity, syntactic complexity, and language sophistication (McNamara, Crossley, & McCarthy, 2010), and such differences may alter the measurement burden imposed on automated scoring configurations.

Treating uniform dependability across responses as an empirical question rather than an assumption is therefore a natural extension of the conditional-precision concern, even though the Standards directly address score levels and subgroups rather than response-space strata.

Building on this response-side/scoring-side distinction, the present study contributes an entropy-conditional generalizability framework for evaluating whether score dependability is uniform across response conditions. The framework has three components; its central methodological distinction—between analytical and empirical estimands of design adequacy—threads through all three. First, it conceptualizes automated scoring configurations as a universe of admissible measurement conditions rather than as incidental modeling details. In the present illustration, configurations are defined by combinations of encoder architectures and scoring-head families within a fixed development pipeline; in other pipelines, additional implementation choices—random seeds, prompts, calibration procedures, or model versions—may enter the admissible set. Second, it conditions G- and D-study evidence on response-level entropy strata, treating entropy as an operational stratification variable. This step allows aggregate dependability estimates to be compared with condition-specific estimates, thereby identifying response conditions for which a nominally adequate scoring design may be insufficient or inefficient. Third, the framework links analytical D-study projections with empirical configuration sweeps over a finite scoring pool. This comparison distinguishes model-based estimates of design adequacy from the dependability actually attainable under concrete scoring configurations. These two estimands of design adequacy—analytical and empirical—can be cross-validated against one another, and divergence between them carries diagnostic information about the configuration universe. The framework is demonstrated in automated essay scoring because contemporary AES systems make the configuration universe explicit, enumerable, and operationally consequential; however, the same logic applies to measurement settings in which multiple scoring, rating, or observation configurations are available and where reliability may vary across response conditions.

We illustrate the framework with the Cambridge Learner Corpus First Certificate in English (CLC-FCE) dataset, in which each candidate contributes two same-session responses—the repeated-response structure that separates person variance from response- and task-specific variance, and that single-response corpora cannot provide. Twenty-four automated scoring configurations (six text encoders crossed with four scoring-head families), trained in a fixed development pipeline and applied to a 300-candidate holdout excluded from training, define the universe of admissible scoring conditions; the random seed is held constant at training and is not varied as a facet here. Shannon entropy stratifies responses into measurement-design conditions, and an officially designated 2001 sample provides an external check across examination forms. Three research questions follow:

1. To what extent are automated essay scores generalizable and dependable across alternative scoring configurations within a fixed development pipeline?
2. Do relative generalizability and absolute dependability vary across entropy-defined response conditions, and is any variation robust to how strata are constructed and to essay length?
3. What crossed scoring designs are required to reach target levels of dependability—overall and within response strata—how dependable are realistic single-configuration deployments, and do analytical D-study projections agree with empirical configuration sweeps over a finite scoring pool?

The remainder of the paper proceeds as follows. Related Work situates the framework within recent extensions of generalizability theory to AI-mediated measurement, evaluation frameworks for automated scoring, and the psychometric evaluation of model-based scoring. Method defines the measurement model, corpus, scoring-configuration grid, entropy stratification, and estimation and projection procedures. Results report the pooled variance structure, condition-specific dependability, analytical decision-study projections, empirical configuration sweeps, and the external check. The Discussion covers explanations, links to prior literature, and implications for measurement design.

## Related Work

### Generalizability Theory in AI-Mediated Measurement Designs

Generalizability theory provides the error-source accounting required to evaluate measurement designs that include AI-generated conditions. It defines a universe of admissible observations, decomposes observed-score variance into components for the object of measurement, the design facets, and their interactions, and uses decision studies to project relative generalizability and absolute dependability under alternative procedures (Cronbach, Gleser, Nanda, & Rajaratnam, 1972; Brennan, 2001; Webb, Shavelson, & Haertel, 2006). The replication required for a target dependability can thus be derived rather than assumed, and contributions of tasks, scoring conditions, and person-by-condition interactions separated rather than absorbed into a single coefficient (Nunnally & Bernstein, 1994).

A recent extension to AI-mediated measurement is Lee, Kim, and Shin (2026), who developed a generalizability-theory framework for randomly parallel testing (RPT), in which examinees receive different but psychometrically similar item sets generated from templates or AI systems. Their framework estimates conditional standard errors of measurement and related reliability indices under RPT designs that may be crossed, nested, or multivariate. The structural move is to treat AI-generated item variants as a sampled realization from a universe of admissible items rather than as a fixed set of forms—evaluating precision locally rather than globally.

The present study addresses an analogous problem on the scoring side. The response is fixed, but the automated scoring system is one realization sampled from a universe of admissible configurations: combinations of encoder architectures and scoring-head families defensible within a fixed development pipeline. (Other pipelines could define the admissible set using additional implementation choices, such as random seeds, prompts, calibration procedures, or model versions; in the present illustration, these are held constant.) Where Lee et al. (2026) ask how to estimate dependability when AI participates in item generation, this study asks how to estimate dependability when AI participates in scoring. The shared logic is analogous, not identical: in both, AI introduces newly replicable conditions into the measurement procedure, and precision should be evaluated conditionally rather than assumed uniform. The estimation strategies differ accordingly. Lee et al.'s framework develops CSEMs under crossed and nested item designs; the present framework conditions G- and D-study evidence on response-level strata and combines analytical D-study projections with empirical configuration sweeps over a finite scoring pool.

Earlier applications of generalizability theory in writing assessment established that dependability is a property of the design and population, not of the instrument alone. Huang (2008) analyzed adjudicated essay scores from three years of a provincial examination and found the variance structure differed systematically between ESL and native-English-speaking examinees, with significantly larger residual error and lower generalizability for ESL students.

Wilson, Chen, Sandbank, and Hebert (2019) showed in decision studies of automated scoring that the prompts required for dependable use depended jointly on the intended decision and the population: .90 reliability required two prompts per genre for nonstruggling writers but four to five for struggling writers; .80 was met with one and two, respectively. Chen, Hebert, and Wilson (2022) extended this with a multivariate analysis of human and PEG scores, finding both reliable overall but differing in relative precision by subgroup—automated scoring more reliable for nonstruggling writers, hand-scoring more reliable for struggling writers. In all three studies, the varied facets concern prompts, genres, populations, or scoring methods, while the internal configuration space of the automated scoring system is held fixed. The present framework adds the scoring side itself to the generalizability design by treating admissible scoring configurations as replicable measurement conditions.

**Automated Scoring as a Measurement-Design Problem**

Frameworks for the operational evaluation of automated scoring have long required evidence beyond agreement with human raters. Williamson, Xi, and Breyer (2012) specified that defensible operational use rests on the fit between scoring capability and assessment purpose, agreement with human scores, associations with independent measures, generalizability across tasks and forms, and consequences for subgroups. Ramineni and Williamson (2013) documented how these requirements are implemented in operational practice. Both treat the automated system as one component of a measurement procedure rather than a standalone prediction model. What they specify, however, is the categories of evidence required—not the variance-decomposition machinery for quantifying how much each element of the scoring design contributes, or for projecting how dependability would change under alternative designs.

Because the scoring configuration is itself a score-generating condition—reflecting choices of text encoder, scoring model, training data, and other implementation choices a pipeline treats as replicable—two configurations built from the same pipeline can assign the same response different scores. The configuration occupies the structural position raters occupy in classical human-scored designs: a replicable source of measurement error whose contribution to score variance, and whose required replication for a given target, can be derived from a variance decomposition. Treating the configuration as a measurement facet—rather than as a fixed implementation detail—is what allows the operational evaluation question to be reformulated as a generalizability-theory question about the universe of admissible scoring conditions.

**Psychometric Evaluation of LLM-Based Scoring**

Recent studies have begun to evaluate generative scoring systems with the same psychometric machinery. Li, Huang, Wu, and Whipple (2024) used G-theory to compare ChatGPT 3.5 and 4 with four college English teachers on 30 CET-4 essays, finding version 3.5 consistently less reliable than the teachers and version 4 consistently more reliable. Gao, Hashim, and Md Yunus (2025) found the same pattern for DeepSeek on 92 CET-4 essays (V3 below the teachers, R1 above). At larger scale, Wang, Huang, Du, Guo, Liu, and Wang (2025) combined G-theory with many-facet Rasch modeling across 4,315 essays in three genres, finding LLM raters adequate for relative ranking but less reliable for absolute standard judgments, with scoring stability differing by model family and rubric structure conditioning the human–machine comparison. Huang and Wilson (2025) further found that prompt design affected both agreement with human scores and comparability across student groups.

Taken together, this literature shows that generative scoring reliability depends on model choice (Gao et al., 2025; Li et al., 2024; Wang et al., 2025), task and rubric structure (Wang et al., 2025), and prompting strategy (Huang & Wilson, 2025). Two limits remain. First, these studies treat models or versions as a small set of named raters; they do not define a universe of admissible configurations within a fixed development pipeline or ask how much an examinee's score depends on which configuration is selected. Second, apart from Wang et al.'s relative–absolute asymmetry—which anticipates, at the rater-behavior level, the distinction between relative generalizability and absolute dependability that the present study estimates at the level of scoring designs—dependability is reported in the aggregate, with no conditioning on properties of the responses scored.

**Response-Conditioned Dependability**

The measurement standards make the conditioning principle explicit. Standard 2.14 requires that conditional standard errors of measurement be reported at several score levels "unless there is evidence that the standard error is constant across score levels," with the rationale documented when constancy is assumed (AERA, APA, & NCME, 2014, p. 46). Standard 2.15 extends the expectation across populations: when credible evidence suggests that conditional standard errors "will differ substantially for various subgroups, investigation of the extent and impact of such differences should be undertaken and reported" (p. 46). Conditioning dependability on strata of the response space motivates an analogous question through a generalizability lens—whether uniform precision across the essays scored should be treated as an empirical question rather than an assumption—although the Standards directly address score levels and subgroups rather than response-space strata.

Operational studies of subgroup behavior in automated scoring have largely addressed mean comparability rather than dependability comparability. Bridgeman, Trapani, and Attali (2012) compared human and machine scores across gender, ethnicity, and country groups in two high-stakes programs, finding them similar for most subgroups, with exceptions that motivated policies limiting machine-score influence on reported results. Amorim, Cançado, and Veloso (2018) showed that systems trained on human ratings can inherit human rater bias. These findings concern the location of scores. Comparable means, however, do not imply comparable precision: a system can behave similarly on average across groups or response types yet yield systematically less dependable scores in some strata, if variance components or person-by-condition interactions differ across them.

For conditional analysis to be substantively meaningful, responses must differ in ways that could plausibly alter the measurement situation. McNamara, Crossley, and McCarthy (2010) showed that essays judged higher in quality were characterized by greater syntactic complexity, greater lexical diversity, and lower word frequency—features of sophisticated language—whereas none of 26 cohesion indices distinguished quality levels. Their question was what predicts quality judgments, not how precisely scores can be assigned; that essays with different linguistic profiles may also present different measurement conditions to a scoring system is the hypothesis investigated here, not a finding of that literature.

Shannon (1948) entropy is used here as an operational stratification variable for evaluating whether variance components and D-study conclusions are stable across response conditions, not as a construct claim about writing quality. As the average information, or surprisal, of a response's token distribution, it indexes lexical unpredictability directly, requires no parser or rubric input, and reflects the full frequency distribution rather than a count of types.

Unpredictability is also the property that, in theories of human language processing, has been associated with greater processing cost (Hale, 2001; Levy, 2008) and with the management of information density under channel-capacity pressure in language production (Levy & Jaeger, 2007). Whether analogous mechanisms operate in automated encoders is conjectural. Conditioning dependability on entropy is, in this sense, an analogue of conditioning measurement precision on score level—motivated by the same conditional-precision concern, while recognizing that the Standards directly address score levels and subgroups rather than response-space strata. Whether the resulting stratification pattern depends on cutpoint choice, on essay length, or on alternative entropy estimators is examined directly in the sensitivity analyses; convergence with established complexity indices is reserved for future work (see Limitations).

These strands jointly define the gap. Generalizability studies of automated scoring have varied prompts, genres, and populations while holding the scoring system fixed (Chen et al., 2022; Wilson et al., 2019); psychometric studies of LLM-based scoring have varied model identity but not a development pipeline's configuration space, reporting dependability in the aggregate (Gao et al., 2025; Li et al., 2024; Wang et al., 2025); and the Standards require conditional reporting of precision wherever uniformity cannot be defended (AERA et al., 2014). No prior study, to our knowledge, treats the admissible configurations of an AES development pipeline as measurement facets in a fully crossed repeated-response design, asks whether the resulting dependability is uniform across response strata, and projects what single-configuration deployment would imply. The framework introduced in this paper addresses that gap.

# Method

## Conditional Generalizability Framework

The framework introduced in this study evaluates whether score dependability is uniform across response conditions when a measurement procedure draws on a universe of admissible scoring configurations. It is specified in six steps, abstracted from any particular dataset or scoring system.

**Step 1.** Define the universe of admissible scoring configurations. A scoring configuration is a specification of implementation choices that a study or operational program treats as replicable—encoder architecture, scoring head family, training procedure, random seed, prompt, calibration step, and so on. The admissible universe is the population of configurations a developer might defensibly draw upon while keeping the rest of the pipeline fixed. The configuration occupies the structural position raters occupy in classical human-scored designs.

**Step 2.** Estimate aggregate variance components on the realized grid. A fully crossed G-study—person × task × {chosen configuration facets}—decomposes observed-score variance into components for the object of measurement, the design facets, and their interactions, with the corresponding relative generalizability ($E\rho^2$) and absolute dependability ($\Phi$) coefficients (Brennan, 2001; Cronbach et al., 1972).

**Step 3.** Define response conditions. Responses are partitioned into strata along an operational conditioning variable that may plausibly alter the measurement situation. The choice of conditioning variable is justified theoretically and tested for sensitivity to construction.

**Step 4.** Re-estimate variance components within each response stratum and compare. Stratum-conditional G/D-study evidence is computed from the same procedure as the aggregate evidence and contrasted with the pooled estimates. The comparison answers whether the design burden is approximately homogeneous across the conditioning axis.

**Step 5.** Project analytical D-study minimum designs. From the pooled and stratum-conditional variance components, decision studies project the crossed designs required to reach conventional dependability targets, including extrapolations beyond the realized facet levels permitted by the random-facet logic.

**Step 6.** Conduct empirical configuration sweeps over the finite scoring pool. The realized grid is enumerated over its admissible sub-designs to ask which concrete configurations actually attain a given target, irrespective of what the analytical projection extrapolates. The two estimands of design adequacy—the analytical projection in Step 5 and the empirical sweep in Step 6—are compared; convergence supports the analytical projection, while divergence carries diagnostic information about the configuration universe (e.g., a homogeneous configuration pool may underrealize an extrapolated target by lacking the variety the analytical components assume).

The Method that follows instantiates this framework on the CLC-FCE corpus.

## Data Source and Analytic Sample

Data were drawn from the publicly released Cambridge Learner Corpus First Certificate in English (CLC-FCE) dataset (Yannakoudakis, Briscoe, & Medlock, 2011), which contains 1,244 anonymized examination scripts from the FCE writing component. Each script is one candidate in a single session and, for 1,237 of the 1,244 scripts, contains two written responses: a required first task (answer 1) and a candidate-selected second task (answer 2). Each response carries an examiner-assigned ordinal mark as a band-and-level code (e.g., 4.1); we mapped the sixteen valid codes (0 and 1.1 through 5.3) onto ordered class scores 0–15, used as the ordinal target for fitting and the common 0–15 reporting metric. Variance-component analyses are read on this reporting scale and do not require treating the original labels as equal-interval. Figure 1 summarizes the data structure, scoring conditions, and analysis workflow.

**Figure 1**

*Measurement Design and Analysis Workflow*

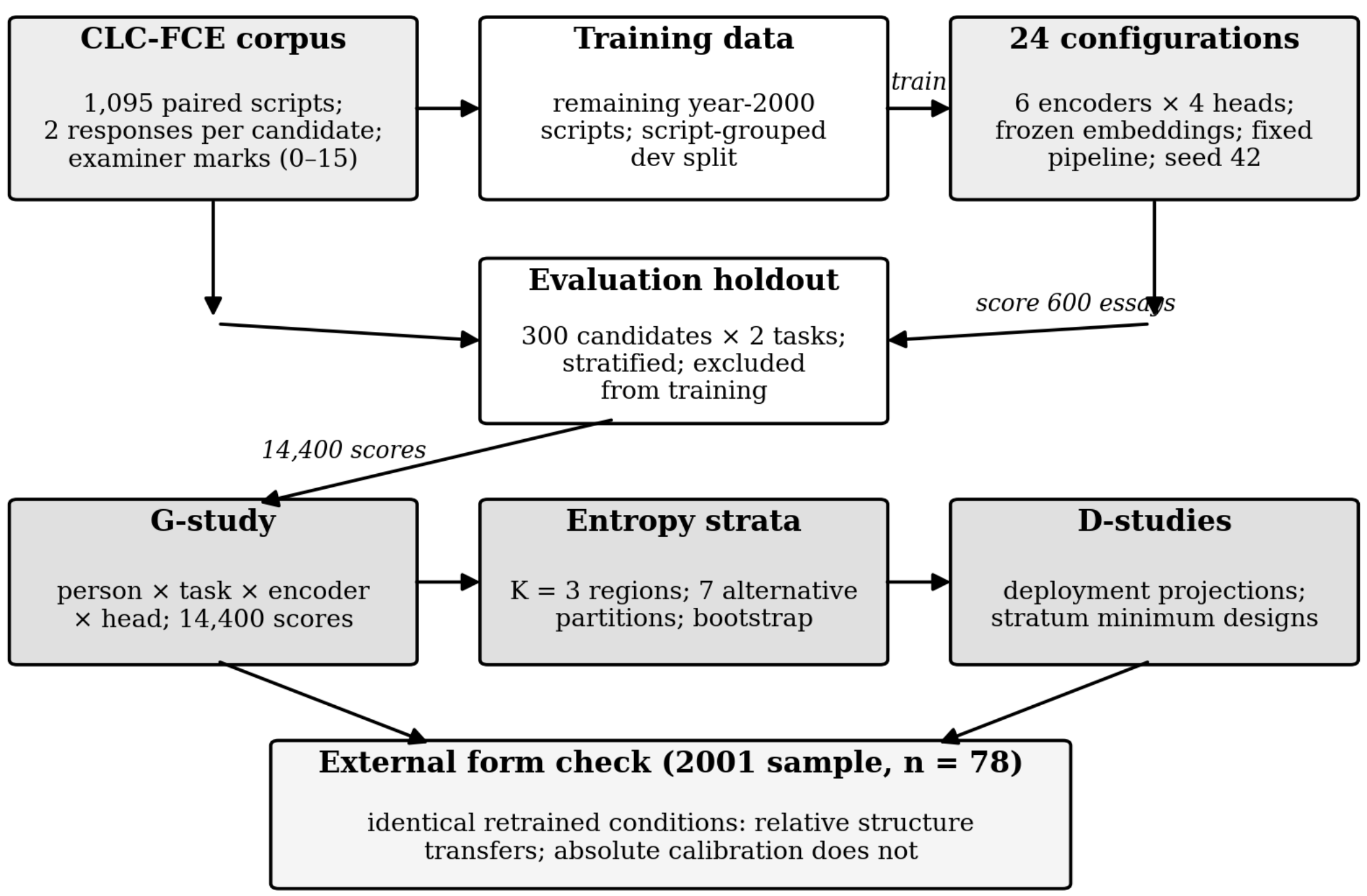

*Note.* CLC-FCE = Cambridge Learner Corpus First Certificate in English.

Response texts were extracted as the learner-original text: the corpus encodes both original learner language and examiner corrections, and we retained the original forms, excluding corrections. Scripts were retained when both responses carried valid ordinal marks (codes 0–5.3; nonstandard codes such as compound or suspended marks were excluded), yielding 1,095 paired scripts: 1,011 from the year-2000 training cohort and 84 from year-2001 forms, including 78 official paired test scripts and 6 additional paired scripts from scattered forms.

The primary evaluation sample was a demographic-stratified holdout of 300 paired scripts (600 essays) drawn from the 1,011 year-2000 paired scripts. Strata were defined by first language crossed with collapsed age band, retaining at least one script from each of the 67 nonempty strata. The officially designated 2001 test sample (only 78 paired candidates) is too small for stable variance-component estimation in the fully crossed design, particularly after conditioning, and was therefore reserved for external robustness checks. The 300 holdout scripts were excluded from all scoring-model training; the remaining 711 year-2000 paired scripts and additional single-response scripts were used for training, with an internal script-grouped split for diagnostics. No script contributed to both training and evaluation, and the absence of overlap was verified computationally for all scoring conditions. On the holdout sample, examiner marks had a mean of 9.36 and a standard deviation of 2.71 on the 0–15 scale.

**Object of Measurement, Universe of Generalization, and Facets**

The object of measurement is the candidate, observed through repeated writing responses on the examination's writing component. What is studied is the dependability of automated scoring procedures, read from the facet variance components: encoder architectures and scoring-head families occupy the structural position raters occupy in classical generalizability designs. Because each candidate contributes two responses, person variance is separated from essay-specific (person-by-task) variance—a separation single-response corpora cannot support.

The task facet was treated as random, with the universe of generalization restricted to FCE-type timed writing tasks as operationalized in the writing component; no generalization beyond it is claimed. With only two task positions, the task main-effect variance carries a single degree of freedom and should be read cautiously; the focus is on person-by-task and person-by-scoring-condition interactions, estimated more precisely. An alternative interpretation treating the two positions as fixed is reported in the supplement, under which person-by-task variance enters universe-score variance and the coefficients are conservative lower bounds (Brennan, 2001, secs. 4.4, pp. 126–128). Because the second task is candidate-selected, prompt identity is not randomly assigned and was used only diagnostically (see Limitations).

The universe of admissible observations comprised the admissible AES configurations within a fixed development pipeline: systems trained on the same data, with the same frozen-encoder procedure and a fixed seed, varying in encoder architecture and scoring-head family. The realized 6 × 4 grid was specified a priori to span encoder scale tiers and the major ordinal scoring-head families. The G-study thus quantifies development-stage scoring-design uncertainty: how much an examinee's score depends on which admissible configuration a developer might defensibly select. Configurations of heterogeneous quality are legitimate members of this universe; robustness to the weakest is established in sensitivity analyses. Operational deployment commits to a single configuration, addressed through D-study projections and empirical configuration sweeps.

A deployed scoring system is a frozen realization of the training procedure, so training stochasticity belongs to development-stage uncertainty about the pipeline, not to the measurement procedure's error structure. The random seed was therefore fixed (at 42) for the present illustration, each encoder-by-head combination treated as a single realized scoring condition, and fold excluded as a facet because cross-validation partitioning is a research-evaluation artifact rather than an admissible operational scoring condition. In other pipelines, random seeds, prompt designs, calibration procedures, or model versions could enter the admissible configuration set as additional implementation choices (see Discussion).

The scoring-head facet warrants further clarification. The realized grid enumerates four head families—cumulative-link ordinal logistic regression, rank-consistent ordinal classification, a pairwise preference-ranking hybrid, and direct linear regression—chosen a priori to span the major approaches to ordinal score prediction within this development pipeline. We adopt a dual stance on the head facet's universe of generalization. The empirical configuration sweep restricts inferences to head families realized in this grid and makes no exchangeability claim beyond it; analytical D-study projections that vary the number of heads obey the random-facet logic and may extrapolate beyond the realized four, treating the realized heads as exchangeable representatives of a broader admissible head pool. These yield two distinct estimands of design adequacy, addressed in Decision-Study Projections and Empirical Configuration Sweeps below and revisited in Limitations.

**Configuration Universe: AES Scoring Conditions**

Each of the 600 holdout essays was scored by 24 AES configurations crossing six text encoders with four scoring-head families. The encoders—MiniLM-L6 (Wang et al., 2020), ELECTRA-small (Clark, Luong, Le, & Manning, 2020), BERT-base and BERT-large (Devlin, Chang, Lee, & Toutanova, 2019), RoBERTa-base (Liu et al., 2019), and DeBERTa-v3-large (He, Gao, & Chen, 2023)—span small, medium, and large tiers across several architecture families. Encoders were frozen: each essay was embedded once per encoder, and only the scoring heads were fit. The four head families span the major approaches to ordinal score prediction: cumulative-link ordinal logistic regression (McCullagh, 1980), rank-consistent ordinal classification (CORN; Shi, Cao, & Raschka, 2023), a pairwise preference-ranking hybrid

—the original benchmark on this dataset (Yannakoudakis et al., 2011)—and direct linear regression. For the ordinal logistic head, continuous scores were probability-weighted expectations over classes; the others produce continuous predictions directly.

Validation diagnostics for the 24 configurations are reported in Supplement S1: the grid achieved moderate agreement on average, ordinal logistic heads performed best, and one configuration (MiniLM-L6 with the ranking head) produced near-constant predictions, retained as a legitimate but weak member of the configuration universe in the sensitivity analyses.

**Generalizability-Study Design and Estimation**

The primary G-study was a fully crossed, balanced person × task × encoder × scoring-head design: 300 persons × 2 task positions × 6 encoders × 4 scoring heads, yielding 14,400 observed scores with one observation per cell. Variance components were estimated under the random-effects model by the expected-mean-squares (EMS) procedure (Brennan, 2001, sec. 3.4). Negative estimates were set to zero before computing summary coefficients, following Brennan (2001, sec. 3.4.6) and Cronbach et al. (1972). Relative generalizability was summarized by the generalizability coefficient ($E\rho^2$) and absolute dependability by the index of dependability ($\Phi$), with universe-score variance defined by the person component (Brennan, 2001, secs. 4.1.3–4.1.4). Because the design has one observation per cell, the four-way person × task × encoder × head component is confounded with pure error and treated as residual. Estimates were independently reproduced by a second implementation, agreeing to six decimal places. Figure 2 displays the design and its 15 estimable variance components.

**Figure 2**

*Venn Diagram of the Fully Crossed Person × Task × Encoder × Scoring-Head Random-Effects Design*

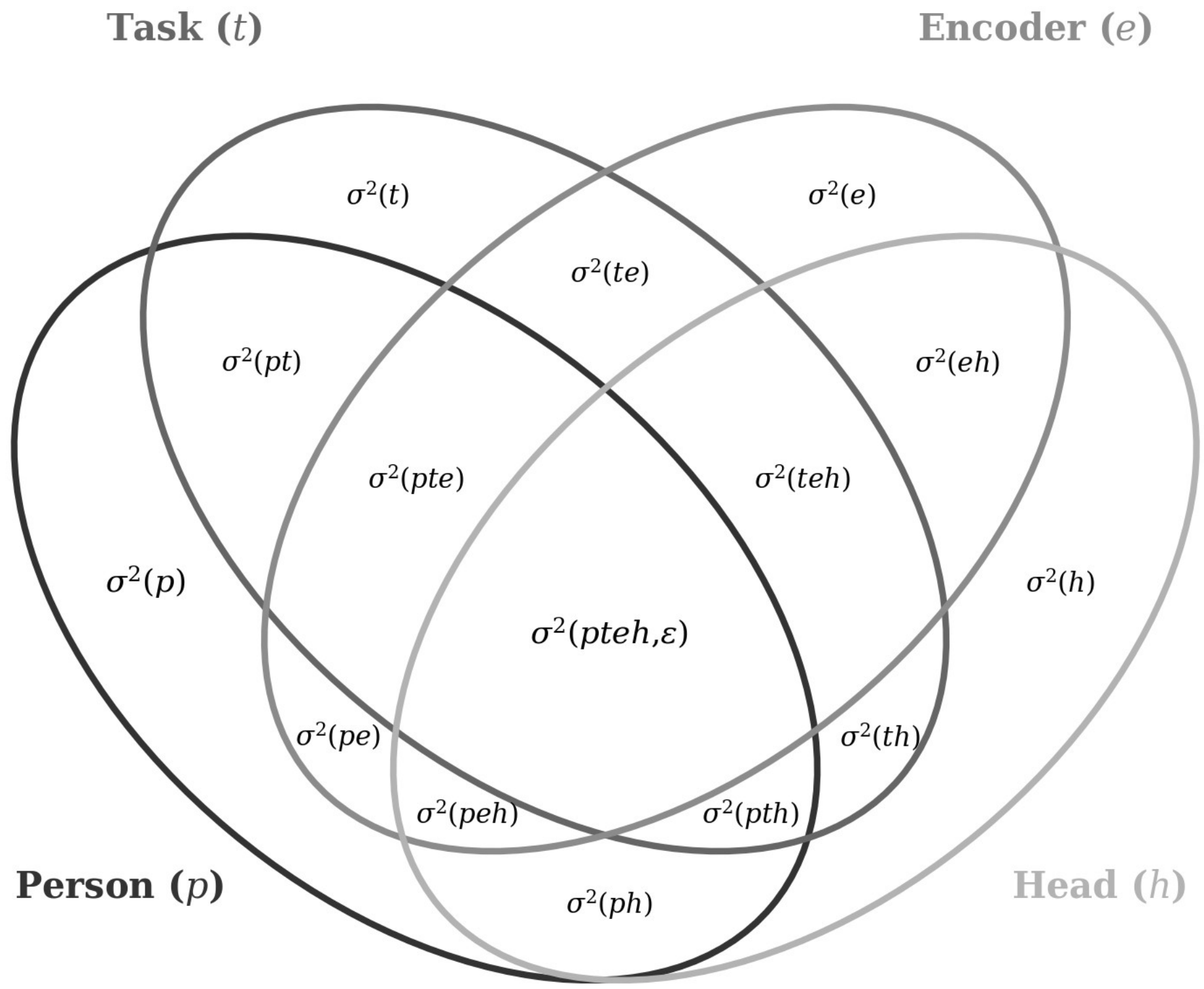

*Note.* The four-way component $\sigma^2$(pteh, ε) is confounded with pure measurement error in the one-observation-per-cell design. Region areas are not proportional to variance.

## Response Strata: Entropy as Operational Conditioning Variable

Shannon entropy was computed for each essay from the empirical distribution of its lowercased word tokens in the learner-original text, in bits (Shannon, 1948). In the holdout sample, essay entropy ranged from 5.52 to 7.13 bits. Entropy served as an operational stratification variable for evaluating whether variance components and D-study conclusions are stable across response conditions.

Two stratifications were used. An equal-variance partition with K = 3 sorted essays by entropy and set cutpoints so that total observed AES-score variance (per essay across the 24 conditions) was balanced across adjacent strata (205, 217, and 178 essays), giving information-balanced strata for stable estimation; because these cutpoints use score variability, the construct-clean inferential check is a pure-entropy tertile partition that uses no score information. The equal-variance cutpoints do not mechanically determine the G-study coefficients, which depend on the universe-score/error decomposition within each stratum. Because stratum membership is an essay-level property—a candidate's two responses can fall in different strata—the stratum-conditional G-studies treated the essay as the object, crossed with encoder and scoring head. A stricter complete-person analysis restricted to candidates whose responses fell in the same stratum is reported as supplementary evidence, read cautiously given the small samples (63, 40, and 44 persons).

Dependence of the stratified results on these construction choices was examined with six alternative partitions that varied K, replaced AES-informed cutpoints with pure entropy tertiles, substituted word-count tertiles, and removed the length confound through residualization and rarefaction (see Supplement S7 for definitions). Uncertainty in stratum contrasts was quantified by an essay-level bootstrap with B = 2,000 within-stratum resamples, a fixed seed, and percentile confidence intervals. Entropy was additionally recomputed with three bias-corrected estimators —Miller–Madow (Miller, 1955), Chao–Shen (Chao & Shen, 2003), and Zhang (2012)—on the full token sequence, to confirm that the stratified pattern does not depend on finite-sample entropy bias (Supplement S8).

**Decision-Study Projections and Empirical Configuration Sweeps**

Two estimands of design adequacy were computed in parallel. Analytical decision studies projected dependability under alternative scoring designs from the pooled nonnegative variance components, varying the number of tasks, encoders, and scoring heads, including single-configuration deployments scoring one or two tasks. Parallel stratum-conditional D-studies projected, within each entropy stratum, the minimum crossed design required to reach conventional targets ($\Phi \geq .70, .75, .80$). These analytical projections obey the random-facet logic and may extrapolate beyond the realized facet levels (e.g., 2 encoders × 5 heads, exceeding the realized four-head ceiling).

An empirical configuration sweep enumerated, within the realized 6 × 4 grid, the dependability attained by every admissible sub-design (combinations of one to six encoders crossed with one to four heads, with the realized task positions). Empirical minima were extracted by selecting the smallest sub-design whose realized $\Phi$ met a target. Because the empirical sweep is constrained to the realized scoring pool, its minima may differ from the analytical projections—particularly where attaining a target would require extrapolation outside the realized facet levels. The two estimands are reported side by side, and divergence is interpreted as diagnostic information about the configuration universe (see Discussion).

Because the four-way component is confounded with pure error, single-condition analytical projections treat the full residual as error and are conservative lower bounds. The empirical sweep does not extrapolate from the full-grid variance components in the same way as the analytical D-study projections; however, because each sub-design is still evaluated from realized scores with one observation per cell, its estimates remain conditional on the same observational structure.

**Robustness Checks**

Three robustness analyses—balanced configuration-quality exclusions (6 × 3, 5 × 4, and 5 × 3 grids), within-task person × encoder × head G-studies for each task position, and an external check retraining all 24 conditions on the officially designated 2001 sample (78 paired candidates)—are reported with the results and in Supplements S2–S5.

## Results

**Preliminary: Validation of the Scoring Grid**

Across the 24 scoring conditions, agreement with examiner marks on the 600 holdout essays was moderate on average (mean MAE = 1.92, mean QWK = .44, mean Pearson r = .56), with wide variation across conditions. Ordinal logistic heads performed best (mean QWK = .63 across encoders; the strongest single conditions reached QWK = .65–.69), and averaging the six ordinal logistic conditions yielded a composite QWK of .67 (r = .73). One condition (MiniLM-L6 with the ranking head) produced near-constant predictions (predicted-score SD = 0.20 against an examiner-mark SD of 2.71; addressed below). The grid thus held substantial ordinal signal alongside real heterogeneity across scoring conditions—the property the measurement-design analyses quantify.

**Aggregate Dependability Across Scoring Configurations**

Table 1 presents the variance components from the fully crossed person × task × encoder × scoring-head G-study. The person component was the largest single source of systematic variance ($\sigma^2$ = 1.158, 31.1% of total), indicating stable examinee signal across conditions. The scoring-facet main effects were negligible: encoder, scoring head, and their two-way interaction together accounted for less than 0.5% of total variance, the task main effect for 0.9%. Error variance was instead dominated by person interactions: the four-way person × task × encoder × head residual (34.1%, confounded with pure error), person × task (9.6%), person × head (7.7%), person × task × encoder (6.3%), and person × task × head (5.3%). Disagreement was thus not a matter of constant offsets; configurations ranked and scored different examinees' responses differently, conditional on the response.

**Table 1**

*Variance Components From the Pooled Person × Task × Encoder × Scoring-Head G-Study*

| Component | Variance | % of total |
|---|---|---|
| Person x task x encoder x scoring head (residual) | 1.271 | 34.1 |
| Person | 1.158 | 31.1 |
| Person x task | 0.360 | 9.6 |
| Person x scoring head | 0.288 | 7.7 |
| Person x task x encoder | 0.237 | 6.3 |
| Person x task x scoring head | 0.197 | 5.3 |
| Person x encoder | 0.087 | 2.3 |
| Person x encoder x scoring head | 0.078 | 2.1 |
| Task | 0.033 | 0.9 |
| Encoder x scoring head | 0.007 | 0.2 |
| Scoring head | 0.005 | 0.1 |
| Encoder | 0.005 | 0.1 |
| Task x scoring head | 0.003 | 0.1 |
| Task x encoder x scoring head | 0.000 | 0.0 |
| Task x encoder | 0.000 | 0.0 |

*Note.* Variance components are shown after negative-component truncation. The four-way person × task × encoder × scoring-head component is confounded with pure error because the design contains one observation per cell. $E\rho^2$ = .773 and $\Phi$ = .763 for the realized 2 × 6 × 4 design.

For the realized 2 × 6 × 4 design, relative generalizability was $E\rho^2$ = .773 and absolute dependability $\Phi$ = .763. The variance-component pattern thus indicates substantial person-level signal in the pooled scoring design alongside nontrivial design-dependent error concentrated in person-by-condition interactions.

**Robustness to configuration quality.** Re-estimating the G-study under three balanced exclusions—dropping the ranking head ($6 \times 3$), the weakest encoder ($5 \times 4$), or both ($5 \times 3$)—raised coefficients modestly ($E\rho^2$ = .79–.80; $\Phi$ = .78–.80) without altering the residual variance share (33.4%–34.3%), confirming that the large person-by-condition interaction is not an artifact of the weakest configurations (see Supplement Table S5).

**Within-task structure.** Separate person × encoder × head G-studies within each task position produced nearly identical dependability (task 1: $E\rho^2$ = .872, $\Phi$ = .871; task 2: $E\rho^2$ = .865, $\Phi$ = .864) and qualitatively the same variance structure: person variance largest (44.0% and 39.6%), the person × encoder × head interaction next (31.6% and 39.0%), person × head exceeding person × encoder (14.1% vs. 9.5%; 12.6% vs. 8.3%), and facet main effects near zero. The two task positions thus behaved as structurally comparable members of the task universe.

**Response-Conditional Dependability Across Entropy Strata**

Table 2 and Figure 3 present the stratum-conditional results. In the information-balanced equal-variance $K = 3$ partition, essay-level dependability was high in all three strata, but absolute dependability declined monotonically with entropy ($\Phi$ = .876, .867, .838 for strata 1–3; $E\rho^2$ = .882, .870, .859). The highest-entropy stratum showed the weakest absolute dependability, a stratum 1 versus stratum 3 difference of $\Delta\Phi$ = .038 (bootstrap 95% CI [.002, .080], $P(\Delta \leq 0)$ = .018)—a modest but substantively relevant local weakening rather than a collapse of dependability in the highest-entropy stratum. Per-stratum bootstrap 95% confidence intervals for $\Phi$ were [.857, .890], [.845, .884], and [.798, .867] for strata 1–3 (Supplement S9).

The decline was robust to how strata were formed, including under the construct-clean pure-entropy tertile partition, which uses entropy rank order alone with no score information ($\Phi$ = .877, .859, .848; $\Delta\Phi$ = .029, 95% CI [−.002, .066], $P(\Delta \leq 0)$ = .035; agreeing with the equal-variance assignment for 95.5% of essays). Under the bias-corrected entropy tertile (Chao–Shen), the same contrast sharpened to $\Delta\Phi = .036$ with its 95% CI excluding zero (Supplement S8). Alternative K = 2 and K = 4 equal-variance partitions likewise placed the weakest absolute dependability in the highest-entropy stratum (K = 2: $\Phi = .877$ vs. .840; K = 4: $\Phi = .868$, .882, .846, .835).

Because raw entropy correlated moderately with essay length ($r = .63$ with word count; $r = .91$ with unique-token count), length was examined directly. Length tertiles alone produced a similar high-end decline ($\Phi = .875$, .872, .845), confirming length is part of the conditioning signal. The pattern nonetheless persisted under three length controls: tertiles of entropy residualized on log word count ($\Phi = .874$, .874, .842; residualized entropy $r = -.01$ with word count; stratum median word counts 188, 181, 184), spline-based residualization ($\Phi = .873$, .874, .843), and rarefied entropy on fixed 95-token subsamples ($\Phi = .870$, .876, .847). Bootstrap stratum 1 versus stratum 3 contrasts excluded zero for both residualized partitions (log: $\Delta\Phi = .033$, 95% CI [.004, .065], $P(\Delta \leq 0) = .0095$; spline: $\Delta\Phi = .031$, 95% CI [.001, .066], $P(\Delta \leq 0) = .022$) and were directionally consistent but weaker for the rarefied check ($\Delta\Phi = .023$, 95% CI [−.006, .055], $P(\Delta \leq 0) = .064$). Under the length-independent partitions the decline concentrated in the highest stratum—strata 1 and 2 nearly indistinguishable—suggesting a threshold rather than a uniform gradient. The local weakening is thus associated with both length and entropy-specific information beyond length, and is not reducible to word count alone. The pattern also persisted under bias-corrected entropy estimators computed on the full text: the highest-entropy stratum remained least dependable under all three estimators, most strongly under Chao–Shen, which reduced the entropy–word-count correlation from $r = .63$ to $r = .23$ (Supplement S8).

**Table 2**

*Stratum-Conditional Dependability Across Entropy-Based Partitions and Sensitivity Checks*

| Partition | n | $E\rho^2$ by stratum | $\Phi$ by stratum | $\Delta\Phi$ | 95% CI | $P(\Delta \leq 0)$ |
|---|---|---|---|---|---|---|
| Equal-variance K = 3 | 205/217/178 | .882/.870/.859 | .876/.867/.838 | .038 | [.002, .080] | .018 |
| Entropy tertiles | 200/200/200 | .884/.861/.866 | .877/.859/.848 | .029 | [-.002, .066] | .035 |
| Equal-variance K = 2 | 310/290 | .880/.858 | .877/.840 | not estimated | not estimated | not estimated |
| Equal-variance K = 4 | 143/167/163/127 | .880/.882 .861/.857 | .868/.882 .846/.835 | not estimated | not estimated | not estimated |
| Length tertiles | 200/200/200 | .877/.872/.860 | .875/.872/.845 | not estimated | not estimated | not estimated |
| Residualized entropy tertiles | 200/200/200 | .877/.875/.855 | .874/.874/.842 | .033 | [.004, .065] | .0095 |
| Spline-residualized tertiles | 200/200/200 | .876/.876/.855 | .873/.874/.843 | .031 | [.001, .066] | .022 |
| Rarefied entropy tertiles | 200/200/200 | .872/.878/.858 | .870/.876/.847 | .023 | [-.006, .055] | .064 |

*Note.* For partitions with bootstrap estimates, $\Delta\Phi$ is the stratum 1 minus highest-stratum contrast based on 2,000 essay-level bootstrap replicates. Values are rounded to three decimals except P = .0095.

Conventional accuracy metrics did not register this pattern. Across the three primary strata, composite MAE was nearly flat (1.63, 1.63, 1.64) and the composite correlation with examiner marks was, if anything, highest in the highest-entropy stratum (r = .68, .69, .72). Average predictive accuracy thus appeared homogeneous across response strata even where absolute dependability was locally weaker—the two families of evidence answer different questions.

A stricter complete-person sensitivity analysis, restricted to the 147 candidates whose two responses fell in the same stratum (n = 63, 40, and 44), produced less stable estimates ($\Phi$ = .712, .681, .752) and is reported in the supplement; given the small per-stratum samples, these carry little evidential weight relative to the essay-level analyses.

**Figure 3**

*Absolute Dependability Across Response Conditions*

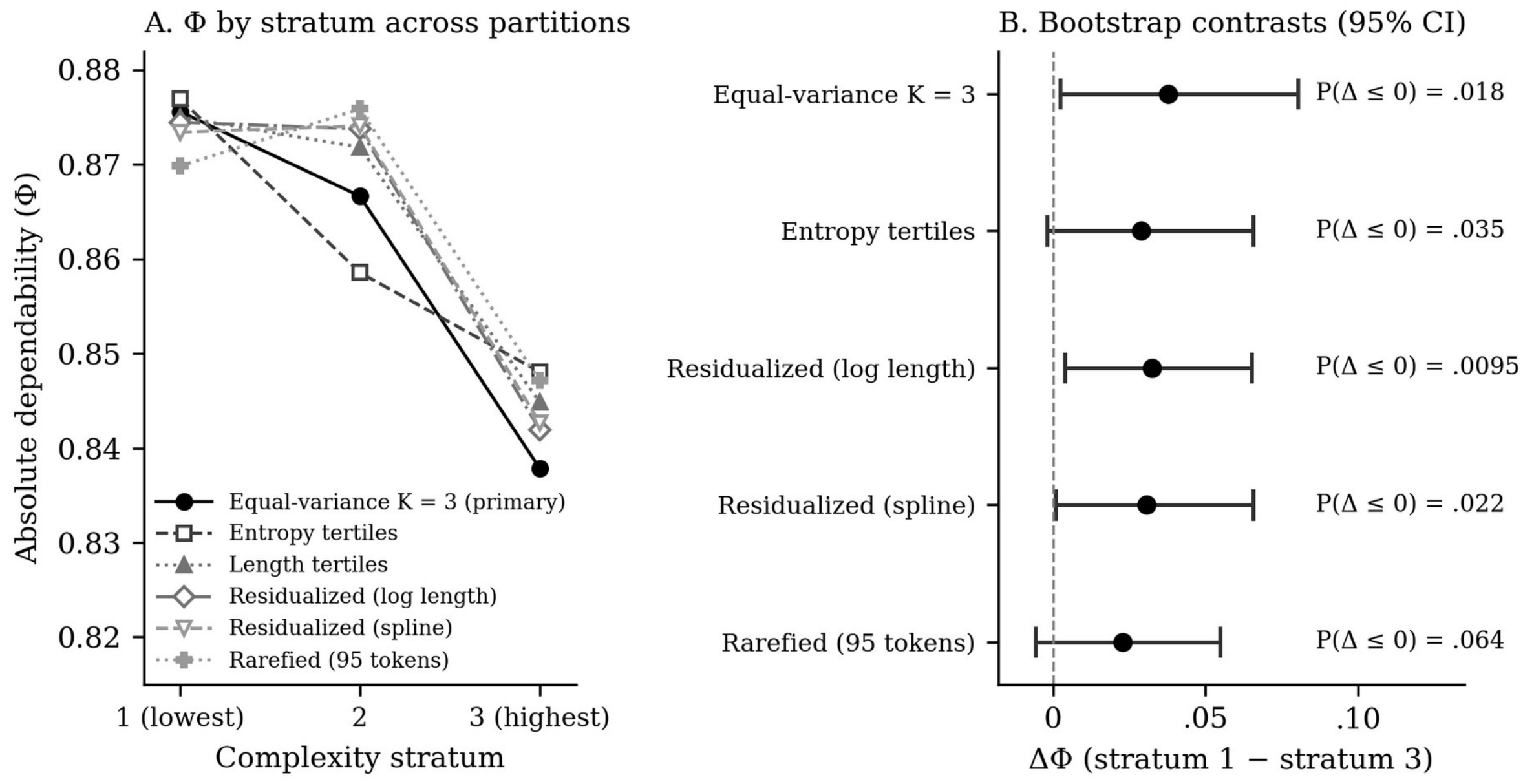

*Note.* Panel A: essay-level Φ by stratum under the primary equal-variance partition and alternative partitions. Panel B: bootstrap contrasts (ΔΦ, stratum 1 minus stratum 3) with 95% percentile confidence intervals (B = 2,000 essay-level resamples within stratum); $P(\Delta \le 0)$ denotes the bootstrap tail probability.

## Analytical D-Study Projections and Empirical Configuration Sweeps

Table 3 and Figure 4 present the two estimands of design adequacy: analytical D-study projections from the pooled variance components, and empirical configuration sweeps within the realized 6 × 4 grid.

**Analytical projections.** Dependability depended sharply on the scoring design. A realistic single configuration (one encoder, one head) scoring both tasks projected Φ = .43 ($E\rho^2$ = .44); with a single task, Φ = .31. Adding conditions on either margin alone was insufficient: six tasks with a single configuration projected Φ = .59, two tasks with a 2 × 2 grid Φ = .63, and a 2-task × 3-encoder × 2-head design remained below .70 (Φ = .665). Approaching the realized-design dependability required expansion on both margins—for example, 4 tasks × 2 encoders × 2 heads projected Φ = .72, still below the realized 2 × 6 × 4 design. Because the four-way component is confounded with pure error, single-configuration projections are conservative lower bounds.

Stratum-conditional analytical D-studies (essay level) projected, for a single configuration, $\Phi$ = .42, .40, and .38 in strata 1–3. Allowing facet extrapolation beyond the realized grid, reaching $\Phi \geq .80$ required 10, 12, and 15 crossed scoring conditions in strata 1–3. For example, the analytical projection for stratum 1 reached $\Phi$ = .803 under a model-based 2-encoder × 5-head design, which is outside the realized four-head grid and therefore not directly attainable in the empirical sweep. Projections that exceed the realized four-head ceiling rest on the random-facet assumption that the realized head families are exchangeable representatives of a broader admissible head pool, and should be read as model-based bounds under that assumption rather than as targets attainable within the present scoring pool (see Limitations).

**Empirical configuration sweeps.** Within the realized 6 × 4 grid, the empirical sweep enumerated every admissible sub-design and identified the smallest whose realized $\Phi$ met each target. The pooled empirical minima rose modestly above the analytical projections at higher targets, and the stratum-conditional empirical minima diverged more visibly from the analytical projections at the highest target: reaching $\Phi \geq .80$ within the realized grid required 12, 12, and 16 crossed scoring conditions in strata 1–3 (three or four encoders crossed with all four heads), and $\Phi \geq .70$ required 6, 6, and 8.

**Divergence between the two estimands.** The empirical minimum-design counts (12, 12, 16 crossed scoring conditions for $\Phi \geq .80$ in strata 1–3) are the primary estimates of design adequacy within the realized scoring pool, because the empirical sweep makes no extrapolation beyond the realized 6 × 4 grid. The analytical projections (10, 12, 15) coincide with the empirical minima in stratum 2 and fall below them in strata 1 and 3, where reaching the target analytically requires head counts (e.g., 2 encoders × 5 heads in stratum 1) that exceed the realized four-head ceiling. The analytical–empirical gap therefore follows mechanically from extrapolating past the realized ceiling under the random-facet assumption for the head family (see Limitations), rather than from a substantive discrepancy in what the two estimands capture; within-ceiling, the two estimands agree (Supplement S11). Within the realized pool, a design sufficient for $\Phi \geq .80$ in strata 1 and 2 (12 empirical conditions) would fall short precisely in the highest-entropy stratum (16 empirical conditions). Aggregate or lower-stratum evidence can thus mischaracterize design adequacy under heterogeneous response conditions.

**Table 3**

*Decision-Study Projections and Empirical Configuration Sweeps*

**Panel A. Selected pooled deployment projections**

| Design | Conditions | $E\rho^2$ | $\Phi$ |
|---|---|---|---|
| 1 task x 1 encoder x 1 head | 1 | .315 | .311 |
| 2 tasks x 1 encoder x 1 head | 1 | .438 | .432 |
| 6 tasks x 1 encoder x 1 head | 1 | .592 | .585 |
| 2 tasks x 2 encoders x 2 heads | 4 | .639 | .631 |
| 2 tasks x 3 encoders x 2 heads | 6 | .674 | .665 |
| 4 tasks x 2 encoders x 2 heads | 4 | .729 | .722 |
| 2 tasks x 6 encoders x 4 heads | 24 | .773 | .763 |

**Panel B. Stratum-conditional empirical minimum designs**

| Stratum | Φ 1x1 | Φ 2x2 | Φ 3x2 | Φ 6x4 | Min .70 | Min .75 | Min .80 |
|---|---|---|---|---|---|---|---|
| 1 | .417 | .680 | .735 | .876 | 2e x 3h = 6 | 2e x 4h = 8 | 3e x 4h = 12 |
| 2 | .401 | .664 | .721 | .867 | 2e x 3h = 6 | 2e x 4h = 8 | 3e x 4h = 12 |
| 3 | .379 | .630 | .683 | .838 | 2e x 4h = 8 | 3e x 4h = 12 | 4e x 4h = 16 |

*Note.* Panel A projections use pooled variance components. Panel B reports the smallest empirical crossed scoring design within the realized 6 × 4 configuration grid; e = encoder and h = scoring head.

**Figure 4**

*Decision-Study Projections of Absolute Dependability*

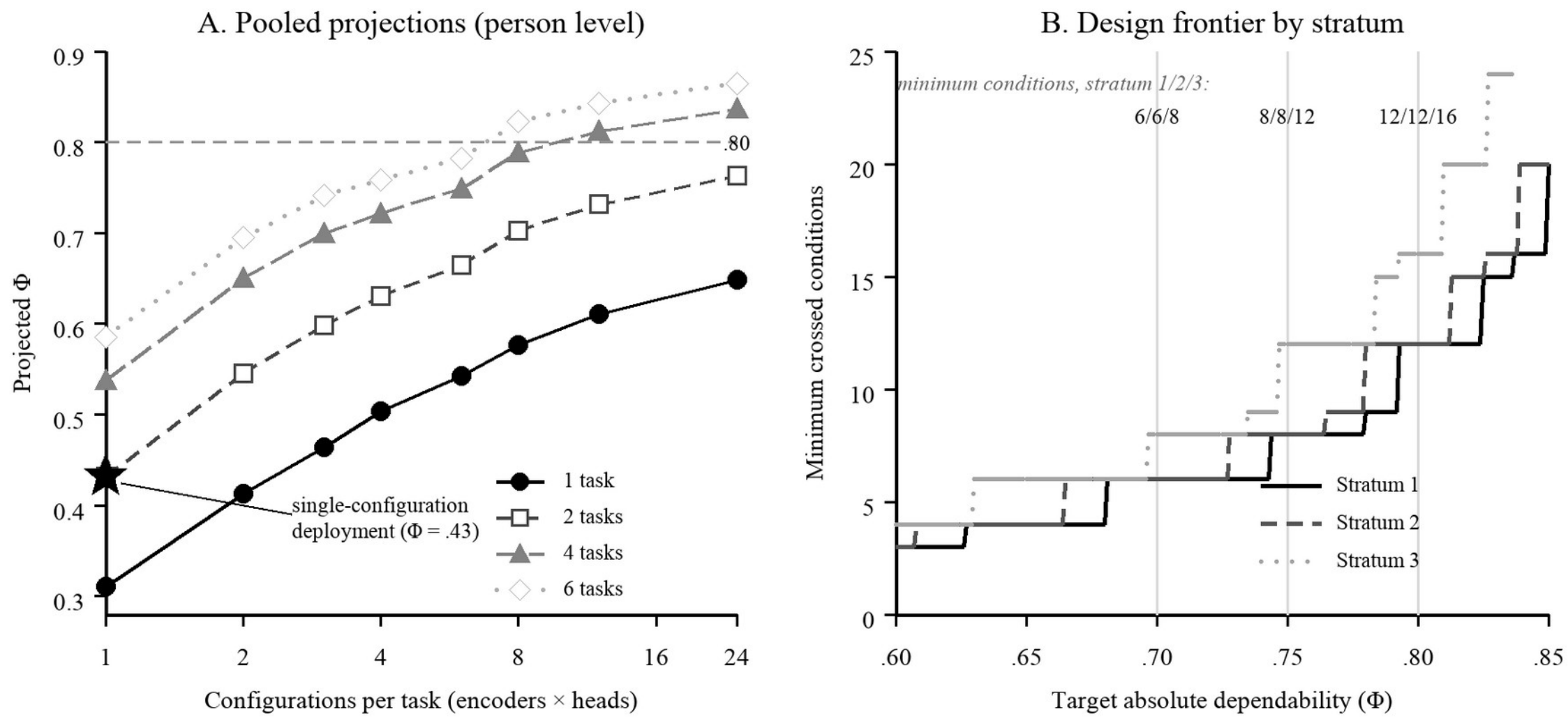

*Note.* Panel A: person-level projections from the pooled variance components; the star marks a single-configuration deployment scoring both tasks. Panel B: the design frontier by stratum, showing the minimum number of crossed scoring conditions required to reach each target Φ. Single-configuration projections treat the full residual as error and are conservative lower bounds.

## External Check on the 2001 Sample

An external check on the officially designated 2001 test sample (78 paired candidates; 3,744 scores from identically retrained conditions) showed that relative generalizability replicated closely ($E\rho^2 = .769$ vs. .773) and person-side variance components were nearly identical. Absolute dependability, however, was substantially lower ($\Phi = .533$ vs. .763), because conditions trained on year-2000 forms shifted their score locations on the 2001 forms by 1.64 points on average—five times the examiner-mark shift—with magnitude varying across scoring-head families (0.86 to 2.34 points). Relative orderings thus transferred, but score calibration did not; both the variance-component comparison and the calibration shifts are reported in Supplement S5 (Table S4 and Figure S4).

# Discussion

## Primary Findings

This study demonstrates that generalizability evidence can change substantively when D-study conclusions are conditioned on response-level uncertainty, and that the design adequacy of an automated scoring procedure is best understood as two distinct estimands—what the pooled variance components imply under unconstrained design extrapolation, and what a finite scoring pool can actually deliver. Four findings carry this argument.

First, on the **configuration side**, scoring-facet main effects were negligible while person-by-condition interactions dominated the error structure. Configurations did not disagree by constant offsets but about particular examinees' particular responses. Treating encoder and scoring-head as measurement facets rather than as fixed implementation details is what makes this variance pattern visible.

Second, on the **response side**, absolute dependability was locally weaker in the highest-entropy stratum ($\Phi = .88 \rightarrow .84$ across entropy strata). The decline was modest, robust to construction, only partly attributable to length, and invisible to conventional accuracy metrics, which were flat across the same strata. The two families of evidence—accuracy metrics versus variance-component dependability—answer different questions about the same scoring system.

Third, the **two estimands of design adequacy** diverged in strata 1 and 3 but agreed in stratum 2. The analytical D-study extrapolates from pooled variance components to a design with sufficient facet variety; the empirical configuration sweep can only construct sub-designs realizable from the 24 configurations actually scored. Where they diverge, the divergence is diagnostic for the configuration universe rather than a flaw in either estimate: the empirical minimum exceeds the analytical projection when the realized pool lacks the configuration variety that an unconstrained design would mobilize.

Fourth, the **external check** on the 2001 sample showed an asymmetry: relative generalizability and person-side variance structure transferred across forms nearly unchanged ($E\rho^2 = .769$ vs. .773), while absolute dependability did not ($\Phi = .533$), because conditions trained on year-2000 forms shifted their score locations on the 2001 forms by different amounts for different scoring heads.

**Plausible Explanations**

These interpretations identify where score variability arises, not why; the mechanisms are plausible accounts, not conclusions the analyses can adjudicate.

The dominance of person-by-condition interactions over facet main effects suggests that encoders and scoring heads place partially different operational emphases on the same responses. Encoders differ in the features their representations preserve; head families differ in the loss structure mapping a representation to the score scale. The pattern is consistent with configurations agreeing about most responses but resolving atypical ones differently.

One plausible account of the entropy gradient is informational, by analogy with theories of human language processing: higher-entropy responses have flatter, less predictable lexical distributions; in human processing such distributions have been associated with greater surprisal-driven processing cost (Hale, 2001; Levy, 2008), and whether analogous mechanisms operate in automated encoders is conjectural. This fits three details—the gradient concentrates in the highest stratum; it persists when length is removed; and it appears in absolute before relative coefficients, as expected if disagreement concerned score location more than ordering. Other contributors these data cannot separate include unmodeled prompt effects, rater behavior on atypical responses, and construct-relevant features correlated with entropy.

The two-estimand divergence in strata 1 and 3 is consistent with a configuration-pool composition explanation: the analytical projection assumes a universe whose facet levels can be replicated as needed, while the realized pool of 24 configurations has a fixed 6-encoder, 4-head ceiling. When stratum-conditional precision requires a configuration set richer in head variety than the realized pool can provide (stratum 1: analytical 2 × 5 vs. empirical maximum 4-head ceiling), the empirical sweep falls back on broader designs to compensate.

The relative–absolute asymmetry in the external check admits a calibration account. The person-side signal that drives relative ordering is shared across forms, and it transferred. Score location is form-specific: configurations trained on year-2000 responses met the 2001 second-task responses at different points of their score scales, and head families—whose objectives calibrate location differently—shifted by different amounts. What failed to transfer was thus not the measurement signal but the score calibration, the component absolute decisions consume and relative ones do not. With two tasks and 78 candidates these form-level components carry a single degree of freedom, so the account is structural, not a precise estimate of form effects.

**Convergence with Prior Literature**

The framework introduced here aligns most directly with Lee, Kim, and Shin (2026)'s generalizability framework for randomly parallel testing. Where Lee et al. develop conditional standard errors of measurement for AI-generated item variants on the response side, the present framework conditions G- and D-study evidence on response-level strata when AI-mediated scoring introduces replicable scoring configurations. The two lines of work share a structural move: when AI participates in the measurement design, precision must be estimated conditionally rather than assumed uniform, and the estimation tools must be adapted to the AI-introduced random facet. They differ in which side carries the newly replicable AI-mediated facet and, consequently, in which estimands are emphasized—CSEMs under crossed and nested item designs on the one hand, conditional $\Phi$ and the analytical–empirical estimand pair on the other.

The findings converge with prior generalizability work in automated scoring while extending its locus. Wilson, Chen, Sandbank, and Hebert (2019) showed the prompts required for dependable automated scores depended on the intended decision and population, and Chen, Hebert, and Wilson (2022) that the relative precision of human and automated scoring differed by subgroup. The present results carry that logic to the scoring system's configuration space: dependable use depends not only on how many responses are collected but on how many—and which—scoring conditions generate them, and on entropy-defined response conditions. Huang's (2008) ESL versus native-speaker difference is a comparable earlier instance, with the conditioning variable moved from examinee population to response properties.

The convergence with recent LLM-rater psychometrics is specific. Wang, Huang, Du, Guo, Liu, and Wang (2025) found LLM raters adequate for relative ranking but less reliable for absolute standard judgments; we find the same asymmetry across scoring configurations—across entropy strata (the gradient appears in $\Phi$ before $E\rho^2$) and across forms (relative structure transferred; calibration did not). Its recurrence across rater types, levels, and samples suggests a structural feature of model-mediated scoring, not an artifact of one system. Li, Huang, Wu, and Whipple (2024) and Gao, Hashim, and Md Yunus (2025) reported reliability differing sharply between versions of one model family; our variance components locate that variability in configuration-by-person interactions rather than main effects, a distinction aggregate comparisons cannot make.

The findings also give empirical content to the measurement standards. Standards 2.14 and 2.15 motivate analogous conditional-precision questions, although the Standards directly address score levels and subgroups rather than response-space strata (AERA et al., 2014, p. 46); the entropy-conditioned results extend the same conditioning logic to regions of the response space. We diverge from common practice on the evidential role of accuracy metrics: agreement statistics were flat across strata in which absolute dependability declined, so the validation evidence customarily reported for AES systems (Ramineni & Williamson, 2013; Williamson et al., 2012) does not substitute for design-level dependability analysis. The form-level calibration shifts echo, at the configuration level, the subgroup exceptions behind score-moderation policies (Bridgeman et al., 2012).

**Implications for Measurement Practice**

For educational measurement practice, the results argue for three implications. First, scoring-design decisions should rest on D-study projections rather than aggregate coefficients: a program adopting a single deployed configuration for absolute decisions cannot infer its dependability from grid-level results, and the projections here suggest such designs may fall well short of conventional thresholds even when the aggregate coefficient looks strong. Second, where responses vary along the conditioning axis, dependability targets should be checked within response strata, since a design sufficient in aggregate may miss the target in response regions with the greatest conditional design burden. Third, design auditing benefits from reporting both estimands—the analytical D-study projection and the empirical configuration sweep within the realized scoring pool—because their divergence diagnoses whether the pool's configuration variety is the limiting factor.

**Limitations and Future Methodological Agenda**

Several limitations bound these conclusions and indicate directions for methodological development. The task facet is the thinnest element of the design: it has only two levels, and the second task is candidate-selected rather than randomly assigned. Task-related variance components therefore carry a single degree of freedom, and the two positions do not represent independent random draws from the universe of FCE-type timed writing tasks to which generalization is restricted. Extending the conditional framework to broader task populations would require designs with more prompts under random or systematic assignment. The configuration universe is one development pipeline (frozen pretrained encoders, head-only fitting, a fixed seed, a $6 \times 4$ grid not randomly sampled); fine-tuned or instruction-based systems would benefit from parallel analyses, as would pipelines in which random seeds, prompts, calibration procedures, or model versions enter the admissible configuration set as additional implementation choices. Configuration quality was heterogeneous; balanced exclusions left the conclusions intact, but different quality screens could yield different aggregate levels.

One assumption load-bearing for the analytical D-study projections deserves separate acknowledgment. Reaching $\Phi \geq .80$ in strata 1 and 3 was projected analytically at 10 and 15 crossed scoring conditions—counts that include head designs (e.g., 2 encoders × 5 heads) exceeding the realized four-head ceiling. These projections treat the four realized head families as exchangeable representatives of a broader admissible head pool, in the sense of generalizability theory's random-facet logic, rather than as a complete enumeration of admissible heads. The empirical configuration sweep within the realized $6 \times 4$ grid (12 and 16 conditions for the same targets) makes no such assumption and is the primary design-adequacy estimate within the present pool; the analytical projection is the corresponding bound under head exchangeability, and the two estimands agree within the realized ceiling (Supplement S11). Whether the head family is better modeled as a finite enumeration or as a sample from a richer architecture space is an empirical question this design cannot adjudicate. Larger and more heterogeneous head libraries—including LLM-as-judge variants, multi-prompt evaluation frameworks, and instruction-tuned scoring heads—would provide direct evidence on the exchangeability assumption and the corresponding boundary between the two estimands.

Entropy is one operational conditioning variable; the framework accommodates alternatives (e.g., bias-corrected entropy estimators, mature complexity indices, or rubric-derived strata), and convergent-validity studies relating entropy to such alternatives would strengthen the conditioning rationale. Although the gradient survived length controls, the strata should be read as response regions rather than a pure lexical-diversity dimension. The equal-variance partition is diagnostic rather than confirmatory, mitigated by the AES-free tertile replication. Examiner marks are single ratings containing rater error this design cannot separate. Finally, all results derive from one examination program and one learner population on a 0–15 scale, and the external sample, while official, is small.

The framework opens three methodological extensions worth pursuing. First, the divergence between analytical and empirical estimands invites theoretical work on when the two estimands should be expected to converge and when their divergence is informative—plausibly a function of configuration-pool entropy itself. Second, simulation studies could derive sampling distributions for conditional $\Phi$ under heterogeneous configuration universes, providing analytical complements to the bootstrap CIs reported here. Third, the same workflow can be applied to other measurement designs in which AI generates a sampled facet (e.g., rater × case medical assessments under AI-generated cases, multi-source classroom observation under AI-mediated rubric application), which would test whether the analytical–empirical divergence patterns documented here generalize beyond automated essay scoring.

## Conclusion

This study illustrates a conditional generalizability framework for evaluating whether score dependability is uniform across response conditions when AI participates in the scoring design. The framework asks two questions a single aggregate coefficient cannot answer: whether dependability holds across the response strata a scoring procedure encounters, and whether the design adequacy implied by analytical variance components can actually be attained by a finite scoring pool. In the CLC-FCE data the answer was design-dependent in every direction the framework sees—across configurations, entropy strata, forms, and analytical versus empirical estimands. The variance components automated scoring adds—encoders and scoring heads in the structural position of raters—behaved like rater facets, disagreeing most about the response regions where the conditional design burden was greatest in this dataset, and the empirical configuration sweep diverged from the analytical projection precisely where the realized pool lacked the configuration variety the analytical estimand assumed. For operational decisions, the unit of evidence is the scoring design and the configuration pool that realizes it, not the scoring system in isolation. The same framework applies, with adaptation, to any measurement setting in which multiple scoring, rating, or observation configurations are available and where reliability may vary across response conditions.

*When Dependability Is Not Uniform: Entropy-Stratified Generalizability Analysis of Automated Essay Scoring*

## S1. Validation Diagnostics for the 24 Scoring Conditions

Table S1 reports holdout validation metrics (n = 600 essays) for each encoder × scoring-head condition. The grid average was MAE = 1.92, QWK = .44, Pearson r = .56; the six-encoder ordinal-logistic composite reached QWK = .67 (r = .73).

**Table S1**

*Per-Condition Validation Metrics on the Holdout Sample*

| Encoder | Scoring head | MAE | QWK | Pearson r |
|---|---|---|---|---|
| bert_base | corn | 1.801 | 0.45 | 0.598 |
| bert_base | olr | 1.657 | 0.608 | 0.635 |
| bert_base | ranking_hybrid | 1.986 | 0.249 | 0.532 |
| bert_base | regression | 2.253 | 0.471 | 0.471 |
| bert_large | corn | 1.774 | 0.45 | 0.627 |
| bert_large | olr | 1.59 | 0.649 | 0.675 |
| bert_large | ranking_hybrid | 1.938 | 0.284 | 0.549 |
| bert_large | regression | 2.895 | 0.341 | 0.363 |
| deberta_v3_large | corn | 1.517 | 0.651 | 0.718 |
| deberta_v3_large | olr | 1.539 | 0.688 | 0.706 |
| deberta_v3_large | ranking_hybrid | 1.587 | 0.592 | 0.696 |
| deberta_v3_large | regression | 2.706 | 0.374 | 0.391 |
| electra_small | corn | 1.773 | 0.486 | 0.625 |
| electra_small | olr | 1.57 | 0.657 | 0.687 |
| electra_small | ranking_hybrid | 1.993 | 0.24 | 0.493 |
| electra_small | regression | 1.667 | 0.615 | 0.64 |
| minilm_l6 | corn | 2.122 | 0.09 | 0.463 |
| minilm_l6 | olr | 1.783 | 0.519 | 0.585 |
| minilm_l6 | ranking_hybrid | 2.215 | 0.027 | 0.126 |
| minilm_l6 | regression | 1.951 | 0.507 | 0.529 |
| roberta_base | corn | 1.866 | 0.364 | 0.63 |
| roberta_base | olr | 1.505 | 0.672 | 0.711 |
| roberta_base | ranking_hybrid | 2.107 | 0.136 | 0.355 |
| roberta_base | regression | 2.221 | 0.505 | 0.512 |

*Note.* MAE = mean absolute error on the 0–15 reporting scale; QWK = quadratic weighted kappa on predicted classes. The minilm_l6 × ranking_hybrid condition is near-degenerate (predicted-score SD = 0.20).

## S2. Complete-Person Stratum Sensitivity

Because entropy strata are defined at the essay level, a stricter sensitivity analysis retained only candidates whose two responses fell in the same stratum. The resulting samples are small and the estimates unstable; they are reported for completeness.

**Table S2**

*Complete-Person Stratum-Conditional Coefficients*

| Stratum | n persons | $E\rho^2$ | $\Phi$ |
|---|---|---|---|
| 1 | 63 | 0.736 | 0.712 |
| 2 | 40 | 0.687 | 0.681 |
| 3 | 44 | 0.804 | 0.752 |

## S3. Within-Task Analyses

Person × encoder × scoring-head G-studies estimated separately within each task position, for the primary holdout and the external 2001 sample.

**Table S3**

*Within-Task Coefficients and Variance Shares*

| Sample | Task | $E\rho^2$ | $\Phi$ | Person % | p × e × h % | p × head % | p × enc % |
|---|---|---|---|---|---|---|---|
| Holdout | 1 | 0.872 | 0.871 | 44.0 | 31.6 | 14.1 | 9.5 |
| Holdout | 2 | 0.865 | 0.864 | 39.6 | 39.0 | 12.6 | 8.3 |
| 2001 external | 1 | 0.885 | 0.774 | 24.7 | 21.4 | 4.9 | — |
| 2001 external | 2 | 0.865 | 0.859 | 37.4 | 38.0 | 11.0 | — |

*Note.* In the external sample, within-task absolute coefficients reflect form-specific encoder and head main effects; relative coefficients match the holdout closely.

## S4. Alternative Interpretation: Task Positions as Fixed

Treating the two task positions as fixed components of the examination moves person-by-task variance into universe-score variance. Under this interpretation the pooled coefficients are $E\rho^2 = .893$ and $\Phi = .891$, computed from the same nonnegative variance components (universe-score variance = $\sigma^2(p) + \sigma^2(pt)/2$). The random-task coefficients reported in the main text ($E\rho^2 = .773$, $\Phi = .763$) are therefore conservative lower bounds.

## S5. External Check: Full Variance Component Comparison

**Table S4**

*Variance Components, Holdout Versus 2001 External Sample*

| Component | Holdout σ² | % | 2001 σ² | % |
|---|---|---|---|---|
| Person | 1.158 | 31.1 | 1.176 | 18.9 |
| Task (df = 1) | 0.033 | 0.9 | 1.247 | 20.0 |
| Encoder | 0.005 | 0.1 | 0.022 | 0.4 |
| Head | 0.005 | 0.1 | 0.0 | 0.0 |
| Person × task | 0.36 | 9.6 | 0.387 | 6.2 |
| Person × encoder | 0.087 | 2.3 | 0.091 | 1.5 |
| Person × head | 0.288 | 7.7 | 0.265 | 4.2 |
| Task × encoder | 0.0 | 0.0 | 0.112 | 1.8 |
| Task × head | 0.003 | 0.1 | 0.181 | 2.9 |
| Encoder × head | 0.007 | 0.2 | 0.0 | 0.0 |
| Person × task × encoder | 0.237 | 6.3 | 0.298 | 4.8 |
| Person × task × head | 0.197 | 5.3 | 0.15 | 2.4 |
| Person × encoder × head | 0.078 | 2.1 | 0.174 | 2.8 |
| Task × encoder × head | 0.0 | 0.0 | 0.78 | 12.5 |
| Residual (4-way, with error) | 1.271 | 34.1 | 1.343 | 21.6 |

**Figure S4**

*External Form Check on the 2001 Sample*

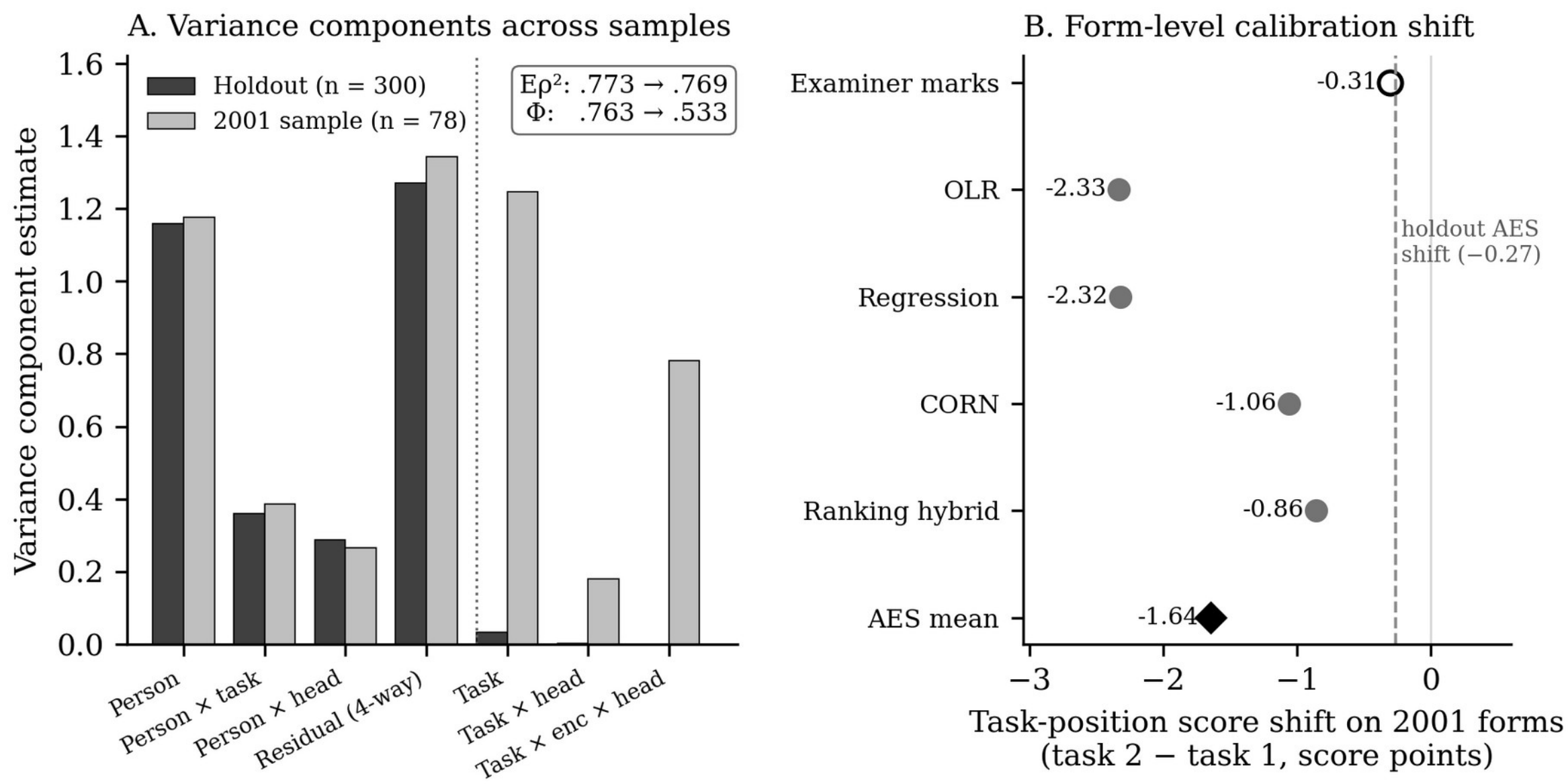

*Note.* Panel A: variance component estimates across samples; task-related components are estimated with a single degree of freedom. Panel B: task-position score shifts on the 2001 forms; examiner marks shifted by 0.31 points between task positions, while scoring-head families shifted by 0.86 to 2.34 points (AES mean: 1.64), against a holdout reference shift of 0.27.

**Table S5**

*Sensitivity of Pooled Dependability to Balanced Configuration-Quality Exclusions*

| Variant | Conditions | $E\rho^2$ | Φ | Person % | Residual % | Stratum Φ (1/2/3) |
|---|---|---|---|---|---|---|
| Full grid | 24 | .773 | .763 | 31.1 | 34.1 | .876/.867/.838 |
| Exclude ranking-hybrid head | 18 | .792 | .786 | 35.7 | 33.4 | .890/.882/.861 |
| Exclude MiniLM-L6 encoder | 20 | .788 | .779 | 33.7 | 34.3 | .889/.876/.849 |
| Exclude both | 15 | .802 | .797 | 37.7 | 34.0 | .898/.883/.863 |

*Note.* Residual % is the four-way person x task x encoder x scoring-head share. Stratum Φ values correspond to the primary equal-variance K = 3 partition.

## S6. Alternative Partition Definitions

Six alternative partitions were constructed to test the sensitivity of the stratum-conditional results to the definition of complexity strata. (a) K = 2 and K = 4 equal-variance partitions applied the same AES-variance-balancing algorithm as the primary K = 3 partition but with different numbers of strata. (b) Pure entropy tertiles (200/200/200) were constructed from entropy rank order alone, with no AES score information; 95.5% of essays received the same stratum label as in the primary partition. (c) Length tertiles were based on word count, exploiting the moderate correlation between entropy and length ($r = .63$ with word count; $r = .91$ with unique-token count). (d) Residualized entropy tertiles used entropy residualized on log word count, yielding a conditioning variable effectively uncorrelated with length ($r = -.01$ with word count; stratum median word counts 188, 181, 184). (e) Spline-residualized entropy tertiles replaced the linear residualization with a spline of log word count ($|r| \leq .01$). (f) Rarefied entropy tertiles used the mean entropy of 500 random 95-token subsamples per essay, which removes length differences constructively ($r = .01$ with word count).

For every partition, uncertainty in stratum contrasts was quantified by an essay-level bootstrap: essays were resampled with replacement within stratum, carrying all 24 condition scores, with B = 2,000 replicates, a fixed resampling seed, and percentile confidence intervals. Variance components and coefficients were re-estimated for each replicate with the same expected-mean-squares procedure and negative-component truncation rule used in the primary analysis.

## S7. Supplementary Figures

**Figure S1**

*Variance Partition of the Fully Crossed Design With Estimated Percentages*

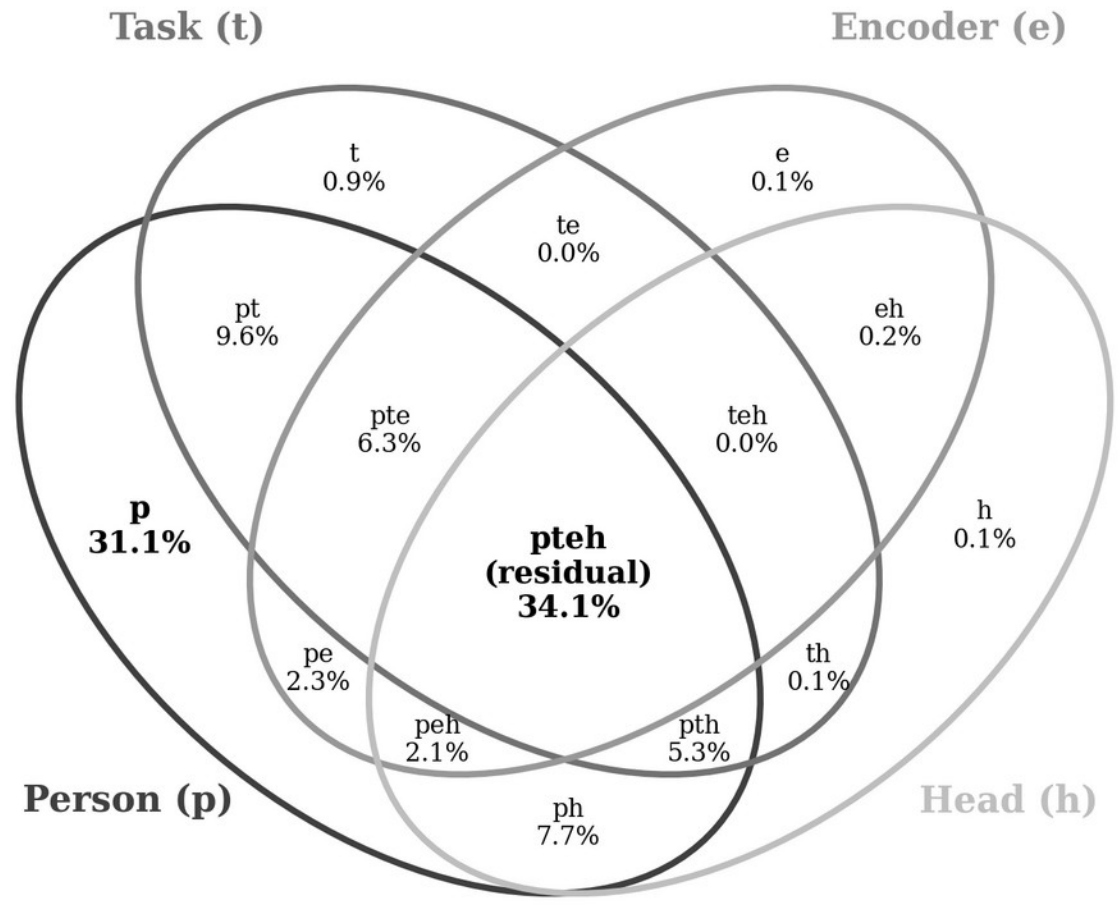

*Note.* Region areas are not proportional to variance. Values are percentages of total variance after truncating negative estimates to zero.

**Figure S2**

*Variance Components as Percentages of Total Variance*

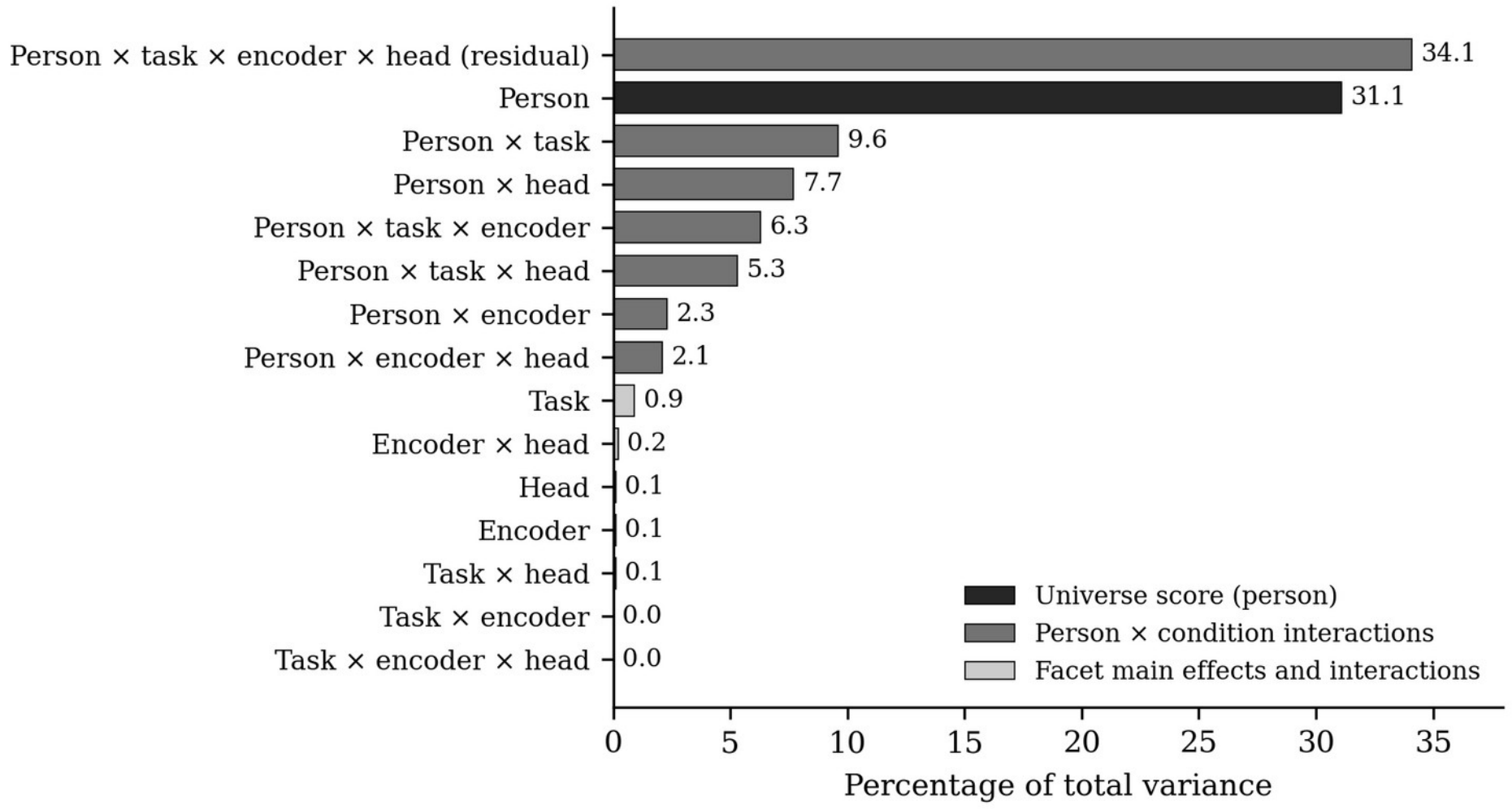

*Note.* Bar-chart rendering of the estimates in Table 1 of the main text.

**Figure S3**

*Conventional Accuracy Metrics and Absolute Dependability Across Strata*

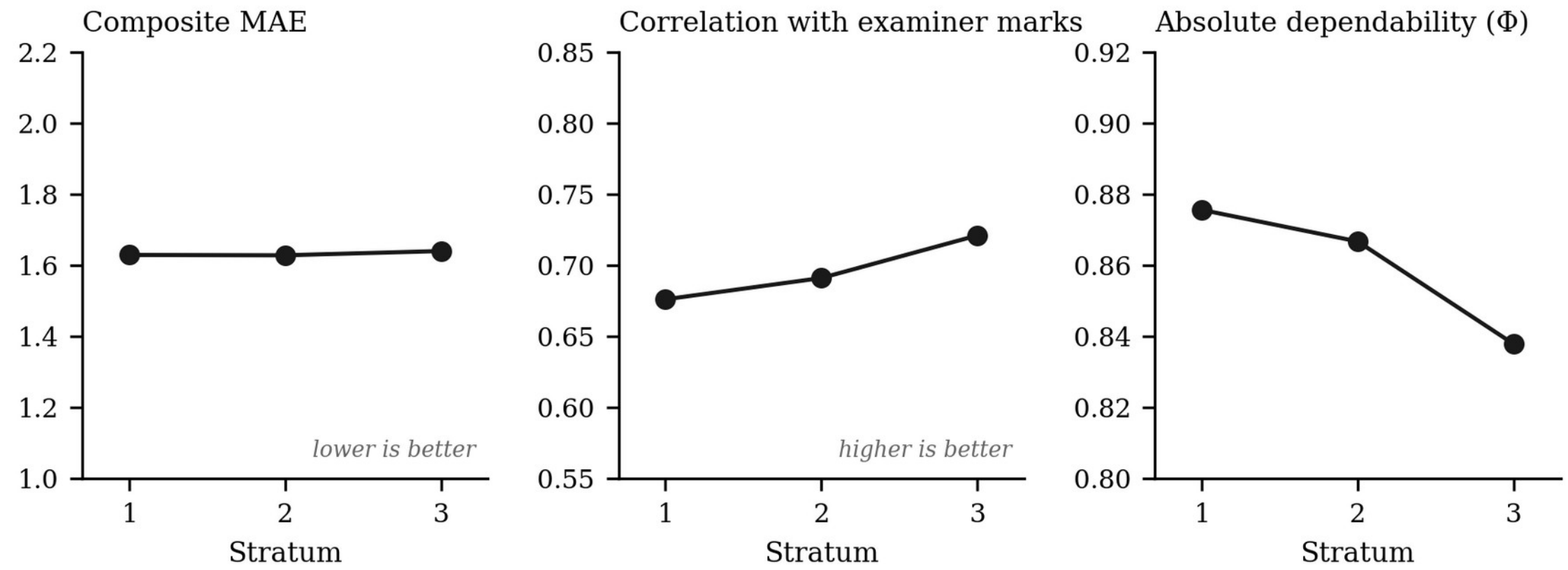

*Note.* Composite MAE and the composite correlation with examiner marks are essentially flat (or improving) across strata, while Φ declines in the highest-complexity stratum.

## S8. Bias-Corrected Entropy Estimators

Because the plug-in (maximum-likelihood) Shannon entropy used for stratification is downward-biased at short essay lengths, and the rarefaction control in the main analysis discards tokens by subsampling to a fixed length, entropy was recomputed for each holdout essay on the full token sequence using three bias-corrected estimators: Miller–Madow (Miller, 1955), Chao–Shen (coverage-adjusted; Chao & Shen, 2003), and Zhang (2012). Essays were then re-stratified (equal-variance and pure-entropy-tertile partitions) and the essay-level stratum-conditional G-study was re-estimated. Across all six corrected analyses the highest-complexity stratum remained the least dependable, and the stratum 1 minus stratum 3 contrast remained positive; the contrast was strongest under Chao–Shen, where its 95% confidence interval excluded zero under both partitions. Under Miller–Madow and Zhang the contrast was directionally consistent but its confidence interval spanned zero, and dependability in strata 1 and 2 was nearly indistinguishable with the decline concentrated at stratum 3—the same threshold pattern reported for the length-controlled partitions in the main text. The corrected estimators also substantially reduced the association between entropy and length (Table S7), so the local weakening is not reducible to essay length or to finite-sample entropy bias.

**Table S6**

*Bias-Corrected Entropy Estimators and Stratum-Conditional Dependability*

| Estimator | Partition | Φ stratum 1 | Φ stratum 2 | Φ stratum 3 | ΔΦ (1−3) | 95% CI | P(Δ≤0) |
|---|---|---|---|---|---|---|---|
| Plain (reference) | Equal-variance | .876 | .867 | .838 | .038 | [.002, .080] | .018 |
| Plain (reference) | Tertile | .877 | .859 | .848 | .029 | [−.002, .066] | .035 |
| Rarefied N=95 (reference) | Tertile | .870 | .876 | .847 | .023 | [−.006, .055] | .064 |
| Miller–Madow | Equal-variance | .870 | .873 | .840 | .030 | [−.005, .071] | .050 |
| Miller–Madow | Tertile | .871 | .875 | .840 | .031 | [−.004, .067] | .039 |
| Chao–Shen | Equal-variance | .875 | .866 | .839 | .035 | [.004, .070] | .013 |
| Chao–Shen | Tertile | .875 | .864 | .839 | .036 | [.004, .071] | .011 |
| Zhang | Equal-variance | .872 | .871 | .842 | .030 | [−.001, .070] | .030 |
| Zhang | Tertile | .872 | .872 | .842 | .030 | [−.003, .066] | .035 |

*Note.* Essay-level stratum-conditional G-studies (object = essay, crossed with encoder and scoring head). ΔΦ is the stratum 1 minus stratum 3 contrast; 95% percentile confidence intervals from B = 2,000 essay-level within-stratum bootstrap resamples (seed = 42). Plain and rarefied rows reproduce the main-text reference values. Φ = index of dependability.

**Table S7**

*Correlations of Bias-Corrected Entropy Estimators With Essay Length and Stratum Agreement*

| Estimator | r with word count | r with unique-token count | Range (bits) | Stratum agreement with plain |
|---|---|---|---|---|
| Plain | .63 | .91 | 5.52–7.13 | 1.00 |
| Miller–Madow | .50 | .85 | 5.91–7.60 | .91 |
| Chao–Shen | .23 | .68 | 6.15–8.13 | .71 |
| Zhang | .48 | .84 | 5.95–7.66 | .90 |

*Note.* Pearson correlations of each essay-level entropy estimate with essay length. Stratum agreement = proportion of essays assigned to the same equal-variance stratum as under plain entropy. Chao–Shen most strongly decouples entropy from length while preserving the dependability gradient (Table S6).

## S9. Empirical Decision-Study Sweep and Bootstrap Intervals

The decision-study projections in the main text are analytical: they extrapolate from the pooled variance components to crossed designs whose facet levels may exceed those realized (e.g., five scoring heads), following standard decision-study logic for random facets. As a complementary audit, dependability was also estimated empirically for every concrete sub-design realizable from the observed 6 × 4 grid, averaging over all subsets of a given size. These two summaries answer different questions and should not be equated. A one-condition design (one encoder, one head, one task) is underidentified, because no non-object facet remains from which to estimate design-dependent error, and is reported as such. Among estimable few-condition designs, concrete sub-designs varied widely—some above and some below the analytical projection—because particular encoder/head subsets include conditions of heterogeneous quality. The empirical audit is therefore bounded at the 24 realized conditions and cannot instantiate the extrapolated designs that the analytical projection recommends; its value is in quantifying the deployment risk of any one realized subset, which reinforces that the scoring design, rather than a single realized system, is the unit of evidence. The analytical minimum designs reported in the main text—12, 12, and 16 crossed conditions for $\Phi \geq .80$ within the realized six-encoder, four-head ranges—were reproduced under an independent reimplementation; extrapolating the facets beyond the realized grid (standard random-facet logic) lowers them to 10, 12, and 15.

**Table S8**

*Analytical Versus Empirical Dependability Across Realizable Deployment Designs*

| Deployment design | Analytical $\Phi$ | Empirical $\Phi$ (mean) | Empirical $\Phi$ (SD) | Empirical $\Phi$ (range) |
|---|---|---|---|---|
| 1 task × 1 encoder × 1 head | .31 | 1.00 (underidentified) | — | — |
| 2 tasks × 1 encoder × 1 head | .43 | .64 | .18 | .17–.85 |
| 2 tasks × 2 encoders × 2 heads | .63 | .57 | .17 | .12–.80 |
| 2 tasks × 3 encoders × 2 heads | .67 | .62 | .12 | .31–.80 |
| Realized 2 × 6 × 4 | .76 | .76 | — | — |

*Note.* Analytical $\Phi$ from the pooled variance components; empirical $\Phi$ averages the index of dependability over all concrete sub-designs of the stated size realizable from the observed 6 × 4 grid. The one-condition design is underidentified (single observation per person).

**Table S9**

*Bootstrap Confidence Intervals for Pooled and Stratum-Conditional Dependability*

| Coefficient | Φ | 95% CI |
|---|---|---|
| Pooled (person × task × encoder × head) | .763 | [.726, .794] |
| Equal-variance stratum 1 | .876 | [.857, .890] |
| Equal-variance stratum 2 | .867 | [.845, .884] |
| Equal-variance stratum 3 | .838 | [.798, .867] |
| Entropy-tertile stratum 1 | .877 | [.859, .892] |
| Entropy-tertile stratum 2 | .859 | [.830, .879] |
| Entropy-tertile stratum 3 | .848 | [.815, .872] |

*Note.* Percentile 95% confidence intervals from B = 2,000 bootstrap resamples (seed = 42); pooled coefficients resampled at the person level (both responses kept together), stratum coefficients at the essay level within stratum.

## S10. Convergent Validity of the Entropy Conditioning Variable

This section reports convergent-validity evidence for Shannon entropy as the operational conditioning variable of the main analysis. Two questions are addressed: (a) whether essay entropy converges with established lexical and syntactic complexity indices, and (b) whether the stratum-conditional dependability gradient observed under entropy strata (Φ = .876, .867, .838 across the equal-variance K = 3 partition) reproduces when the same 600 essays are re-stratified by those established indices. No scoring model was retrained: the 14,400 saved predictions (600 essays × 6 encoders × 4 scoring heads) were only re-stratified, and the equal-variance entropy K = 3 result was reproduced exactly (Φ = .875616, .866663, .837843) before any new index was analyzed.

Lexical-diversity indices (MTLD, MATTR-50, HD-D) were computed with the lexical-diversity package, HD-D via the standard hypergeometric formulation. Lexical-frequency indices were computed directly from the SUBTLEX-US word-form norms (Brysbaert & New, 2009): mean log10(raw count + 1), mean Zipf, and the proportion of tokens below one per million; mean token coverage was 98.25%. Syntactic ratios were computed with a spaCy 3.8.2 dependency-parser approximation (en_core_web_sm 3.8.0) rather than the L2SCA/Stanford Parser stack, which was unavailable; the parser-derived ratios approximate, but are not identical to, the named L2SCA constructs. Chao–Shen bias-corrected entropy was recomputed from the full lowercased word-token sequence. Bootstrap confidence intervals use 2,000 essay-level resamples (percentile method, seed 42); partial correlations control linearly for word count.

Entropy converged most strongly with lexical diversity—Pearson r = .64 (MTLD), .64 (MATTR-50), and .75 (HD-D), rising to .85–.96 after word-count adjustment—and inversely with lexical frequency, while associations with parser-derived syntactic ratios were small and selective. Chao–Shen entropy, which is far less length-dependent (r = .23 with word count vs. .63 for raw entropy), showed the same lexical-diversity pattern even more strongly (Table S10).

**Table S10**

*Convergent Validity: Correlations of Entropy With Established Complexity Indices*

| Index | Construct | r (raw Shannon) [95% CI] | Partial r (word count) | r (Chao–Shen) [95% CI] |
|---|---|---|---|---|
| MTLD | Lexical diversity | .639 [.589, .685] | .851 | .821 [.795, .846] |
| MATTR-50 | Lexical diversity | .638 [.586, .686] | .840 | .765 [.731, .796] |
| HD-D | Lexical diversity | .754 [.712, .792] | .961 | .877 [.857, .895] |
| Mean log SUBTLEX freq. | Lexical frequency | −.283 [−.359, −.203] | −.448 | −.385 [−.454, −.309] |
| Mean Zipf SUBTLEX | Lexical frequency | −.189 [−.265, −.109] | −.322 | −.272 [−.343, −.199] |
| CN/T | Syntactic (parser proxy) | .149 [.056, .239] | .185 | .163 [.070, .253] |
| MLS | Syntactic (parser proxy) | .061 [−.028, .152] | .012 | −.009 [−.094, .084] |
| C/T | Syntactic (parser proxy) | −.011 [−.092, .067] | −.062 | −.037 [−.117, .044] |

*Note.* Pearson correlations with raw Shannon and Chao–Shen entropy; partial r controls linearly for word count. 95% percentile bootstrap CIs (B = 2,000, seed 42), n = 600 essays. Syntactic ratios are spaCy dependency-parser approximations of the named L2SCA constructs. Full Pearson, Spearman, and partial coefficients for all indices and both estimators are in the accompanying workbook.

Re-stratifying the 600 essays by each established index (deterministic 200/200/200 rank tertiles; the same EMS random-effects G-study, negative-component truncation, and bootstrap protocol as the main analysis) gave the pattern in Table S11. The high-end weakening reproduced clearly under MTLD ($\Delta\Phi = .039$, 95% CI [.010, .073]), the lexical-diversity index most aligned with entropy, closely matching the entropy $K = 3$ contrast ($\Delta\Phi = .038$). It was directional but not definitive for sentence length (MLS) and did not reproduce for the clause-per-T-unit proxy (C/T), where the ordering reversed. A frequency-based partition produced a gradient of similar size, but with the higher-frequency tertile least dependable—heterogeneity under a frequency index, not a higher-sophistication-implies-lower-dependability replication. That the gradient appears under lexical-diversity stratification but not under every linguistic partition indicates it is not an artifact of dividing essays into any three groups.

**Table S11**

*Stratum-Conditional Absolute Dependability (Φ) Under Established-Index Re-Stratification*

| Stratifier (ascending tertiles) | Φ str. 1 | Φ str. 2 | Φ str. 3 | ΔΦ (1−3) | 95% CI | P(Δ ≤ 0) |
|---|---|---|---|---|---|---|
| Entropy, equal-variance K = 3 (reference) | .876 | .867 | .838 | .038 | [.002, .080] | .018 |
| MTLD (lexical diversity) | .883 | .861 | .844 | .039 | [.010, .073] | .005 |
| Mean log SUBTLEX frequency | .879 | .864 | .847 | .032 | [.005, .059] | .009 |
| MLS (sentence length) | .883 | .852 | .858 | .024 | [−.004, .054] | .048 |
| C/T (clauses per T-unit, proxy) | .855 | .868 | .878 | −.022 | [−.051, .009] | .920 |

*Note.* ΔΦ is stratum 1 minus stratum 3; 95% percentile bootstrap CIs and P(Δ ≤ 0) use 2,000 essay-level resamples (seed 42). The entropy K = 3 row reproduces the main-text Table 2 reference. Only saved predictions were re-stratified; no model was retrained.

These analyses support describing entropy as a transparent operational proxy for the lexical-diversity and lexical-rarity component of response complexity, not as a general syntactic-complexity measure. The syntactic evidence is convergent at the level of broad dependency-derived ratios rather than a validation against L2SCA output, and replication was strongest for lexical diversity rather than uniform across all established indices.

## S11. Within-Ceiling Analytical–Empirical Convergence

The main text reports two estimands of minimum design adequacy at each dependability target: an analytical D-study projection, which solves variance-component equations for the smallest crossed design meeting the target Φ, and an empirical configuration sweep, which evaluates every combinatorial assignment of encoders and scoring heads over the realized scoring pool. When a target exceeds the single-task ceiling Φ, the analytical projection returns "infeasible" and no comparison is possible. The within-ceiling subset—the nine stratum-by-target cells whose analytical solutions are finite—therefore provides a controlled testbed for assessing analytical–empirical agreement.

Across the nine stratum-by-target cells (strata 1–3 × Φ ≥ .70, .75, .80), the within-ceiling analytical and empirical minimum designs agree exactly: both estimands select the same crossed configuration and the same total number of scoring operations (Table S12). The empirical Φ is slightly lower in every cell because it reflects finite-pool sampling whereas the analytical projection assumes infinite exchangeable levels, but the rank ordering of designs is identical.

The three pooled cells in Table S12 are reported for transparency of the search space but do not converge under within-ceiling matching. Because the analytical projection treats task as a fully crossed random facet and can increase the number of task levels, whereas the empirical sweep is constrained to the single task available in each essay response, the two estimands optimize over different dimensionalities. Within strata, where the task facet is conditioned out, this dimensional mismatch disappears and convergence is exact.

Together, these results indicate that the analytical–empirical divergence at higher dependability targets in strata 1 and 3 (discussed in the main text) is attributable to the finite ceiling of the realized scoring pool, not to a systematic bias in the variance-component algebra. When the comparison is restricted to targets that the pool can plausibly reach, the two estimation routes yield identical design recommendations.

**Table S12**

Within-Ceiling Analytical and Empirical Minimum Designs by Target Φ

| Scope | Target Φ | Within-ceiling analytical design (n; Φ) | Empirical realized design (n; Φ) | Match |
|---|---|---|---|---|
| Pooled | .70 | 4t × 2e × 2h (16; .722) | 1t × 2e × 3h (6; .709) | No† |
| Pooled | .75 | 6t × 2e × 2h (24; .759) | 1t × 2e × 4h (8; .759) | No† |
| Pooled | .80 | 6t × 2e × 3h (36; .800) | 1t × 3e × 4h (12; .811) | No† |
| Stratum 1 | .70 | 2e × 3h (6; .743) | 2e × 3h (6; .724) | Yes |
| Stratum 1 | .75 | 2e × 4h (8; .779) | 2e × 4h (8; .766) | Yes |
| Stratum 1 | .80 | 3e × 4h (12; .825) | 3e × 4h (12; .819) | Yes |
| Stratum 2 | .70 | 2e × 3h (6; .728) | 2e × 3h (6; .702) | Yes |
| Stratum 2 | .75 | 2e × 4h (8; .764) | 2e × 4h (8; .750) | Yes |
| Stratum 2 | .80 | 3e × 4h (12; .812) | 3e × 4h (12; .806) | Yes |
| Stratum 3 | .70 | 2e × 4h (8; .735) | 2e × 4h (8; .724) | Yes |
| Stratum 3 | .75 | 3e × 4h (12; .783) | 3e × 4h (12; .778) | Yes |
| Stratum 3 | .80 | 4e × 4h (16; .810) | 4e × 4h (16; .807) | Yes |

*Note.* Each cell reports the minimum crossed scoring design meeting the target absolute dependability Φ for the indicated scope; n = total scoring operations. † Pooled rows are not within-ceiling matches because the analytical projection can increase task levels whereas the empirical sweep is constrained to one task per response.